\documentclass[10pt,a4paper,logo]{qwenapplication}
\usepackage[numbers,sort]{natbib}

\usepackage{etoolbox}
\patchcmd{\maketitle}{Corresponding author.}{Corresponding authors.}{}{\errmessage{Author-note patch failed}}
\makeatletter
\apptocmd{\ps@firststyle}{\fancyfoot[L]{\footerfont\itshape Contact: \the\correspondingauthor}}{}{\errmessage{Contact-footer patch failed}}
\makeatother
\usepackage{amsmath,amsfonts,bm}

\def\eqref#1{equation~\ref{#1}}

\def\1{\bm{1}}

\DeclareMathAlphabet{\mathsfit}{\encodingdefault}{\sfdefault}{m}{sl}
\SetMathAlphabet{\mathsfit}{bold}{\encodingdefault}{\sfdefault}{bx}{n}

\newcommand{\methodname}{RPMem}
\newcommand{\consolidationgate}{consolidation gate}
\newcommand{\compilationphase}{Single-Session Memory Compilation}
\newcommand{\consolidationphase}{Cross-Session Memory Consolidation}

\newcommand{\compilersft}{Compiler SFT}
\newcommand{\fixedfkl}{Fixed FKL}
\newcommand{\sampledfkl}{Sampled FKL}
\newcommand{\sampledrkl}{Sampled RKL}
\newcommand{\fixedrkl}{Fixed RKL}

\newcommand{\appref}[1]{\mbox{\hyperref[#1]{Appendix~\ref*{#1}}}}

\newcommand{\permabench}{PERMA}
\newcommand{\personamembench}{PersonaMem-v2}
\newcommand{\prefevalbench}{PrefEval}

\newcommand{\coderepourl}{https://github.com/Quark-Medical/rpmem/tree/main}

\newcommand{\permacoreacc}{85.52\%}
\newcommand{\permasevenacc}{86.53\%}
\newcommand{\personamemacc}{40.22\%}

\usepackage{hyperref}
\hypersetup{hypertexnames=false}
\usepackage{url}
\usepackage{enumitem}
\usepackage{booktabs}
\usepackage{amsmath,amssymb}
\usepackage{graphicx}
\usepackage{xcolor}
\usepackage{colortbl}
\usepackage{multirow}
\usepackage{placeins}
\usepackage{float}
\usepackage{wrapfig}
\usepackage{algorithm}
\usepackage{algpseudocode}
\usepackage[skins,breakable]{tcolorbox}
\usepackage{needspace}

\newcommand{\tbd}[1]{#1}

\title{\methodname: Learning Long-Term Recurrent Parametric Memory Across Sessions for LLM Agents}
\author[1,2]{Fanyu Zhao\textsuperscript{*}}
\author[2]{Ruike Cao\textsuperscript{*}}
\author[2]{Liang Dong\textsuperscript{\dag}}
\author[2]{Fugen Yao}
\author[2]{Jian Xu}
\author[2]{Guanjun Jiang}
\author[1]{Yinsheng Li\textsuperscript{\dag}}
\author[1]{Yifei Zhao\textsuperscript{\dag}}
\author[1]{Han Zhang\textsuperscript{\dag}}
\affil[1]{Fudan University}
\affil[2]{Qwen Business Unit of Alibaba}
\correspondingauthor{fyzhao20@fudan.edu.cn}

\begin{abstract}
Long-running LLM agents require memory that persists and evolves across sessions. Text-based memory retrieves and reconstructs past interactions at every query, making long-horizon performance increasingly dependent on retrieval quality and contextual reasoning as histories grow. Parametric memory encodes experience directly into model computation, but existing approaches provide limited support for cross-session memory evolution. Their coupling to a specific backbone further restricts memory reuse after model replacement. We introduce \methodname{}, a two-stage architecture that compiles each session into a model-independent latent memory through forward computation and selectively integrates it with retained memory via a task-trained recurrent gate. The consolidated memory is then mapped to backbone-specific low-rank adaptation (LoRA) parameters, allowing the encoding capability to transfer when the backbone is replaced. Evaluation across three long-term memory benchmarks and five diverse backbones demonstrates broad generalization with near-constant update cost and memory footprint. With Qwen3-8B on \permabench{}, \methodname{} reaches \permacoreacc{}, outperforming the strongest parametric and text-based baselines by 5.32 and 12.98 percentage points, respectively. Ablations validate the complementary roles of session compilation and cross-session consolidation, while dynamics analyses reveal that the gate acquires task-specific memory integration strategies.  These results establish \methodname{} as a lifecycle-independent parametric memory framework that maintains evolving cross-session memory that remains reusable across backbone replacements.
Our code is available at \href{\coderepourl}{\textcolor{blue}{\nolinkurl{https://github.com/Quark-Medical/rpmem/tree/main}}}.
\end{abstract}

\begin{document}
\maketitle

\begingroup
\setlength{\textfloatsep}{12pt minus 6pt}
\setlength{\intextsep}{8pt plus 2pt minus 2pt}
\setlength{\abovecaptionskip}{5pt}
\section{Introduction}

LLM agents increasingly operate across multiple sessions, requiring continuity in both interaction and task execution. Sustained user interaction depends on retaining relevant historical context~\citep{cao2026atpo,zhao2025prefeval,jiang2025personamemv2}. Task execution across episodes similarly requires preserving prior decisions, constraints, and outcomes~\citep{park2023generative,packer2023memgpt,zhong2024memorybank}. Long-term memory has therefore become an essential capability for agent architectures~\citep{hu2025memoryage}.

The dominant approach to long-term memory stores past interactions as external text and retrieves relevant information at query time. Recent work has refined this approach by extracting salient information and consolidating it into compact, retrievable records~\citep{chhikara2025mem0,fang2026lightmem}. Adaptive memory organization further links related entries and revises existing records as new information arrives~\citep{xu2025amem}. Reinforcement learning has also been introduced to optimize memory operations and the selection of retrieved information for downstream tasks~\citep{yan2026memoryr1}. This text-based formulation keeps stored memories inspectable and separate from the backbone. Despite improvements in organization and management, using these memories still requires retrieving relevant records and integrating their contents into the input context at each query. \tbd{Explicitly stored memory remains valuable for questions that require precise evidence and source attribution. Long-running agents, however, may need to repeatedly access information accumulated across sessions. Reducing the inference cost of this repeated access is a central design goal.}


Parametric memory offers an alternative by encoding experience directly into model computation, allowing memory to inform future reasoning without occupying the input context~\citep{charakorn2026doctolora,liu2026shine,tang2026pam,zhang2026metis}. However, deploying parametric memory in long-lived agents requires more than encoding a single context into parameters. Memory updates through forward computation are desirable for scalable deployment, as they avoid the overhead of repeated gradient-based optimization. Methods such as Doc-to-LoRA and SHINE achieve this by mapping bounded text to low-rank adaptation (LoRA) parameters~\citep{hu2022lora} in a single pass~\citep{charakorn2026doctolora,liu2026shine}, but each session produces a static snapshot without a mechanism for incremental cross-session updates.

Selectively retaining, revising, and integrating information across sessions raises a more challenging requirement. MEMORYLLM and M+ maintain updatable memory within the language model~\citep{wang2024memoryllm,wang2025mplus}, while Metis learns forward-updated memory matrices within a memory-augmented backbone~\citep{zhang2026metis}. Yet in existing parametric approaches, memory representations are defined by and coupled to a specific backbone architecture. In deployed systems, the serving model is routinely replaced, and under this coupling accumulated memory cannot be directly reused. Existing adapter-transfer methods support reuse of learned task adaptations across model changes~\citep{wang2024translora,gu2025transpeft,jung2026titok}. Persistent agent memory additionally requires continuity of cross-session updates through those changes. This lifecycle dependence is, we argue, the central unsolved barrier to persistent parametric memory. We define lifecycle independence as preserving and continuing to use accumulated memory across serving-backbone replacements.

We introduce \methodname{} (Recurrent Parametric Memory), an architecture that separates persistent memory from the serving backbone, as illustrated in \autoref{fig:framework}. Training proceeds in two stages. Single-session compilation learns to encode context into a shared latent memory space and decode it into backbone-specific LoRA parameters. Cross-session consolidation trains a lightweight recurrent gate under task supervision to integrate session memories into a fixed-size state. At inference, this state is updated through forward computation and decoded into LoRA parameters for the frozen backbone. Backbone replacement requires adapting only the decoder, preserving the accumulated memory and recurrent updater so that memory remains usable and continues evolving. Our contributions are summarized below:


\begin{figure}[t]
    \centering
    \includegraphics[width=0.96\linewidth]{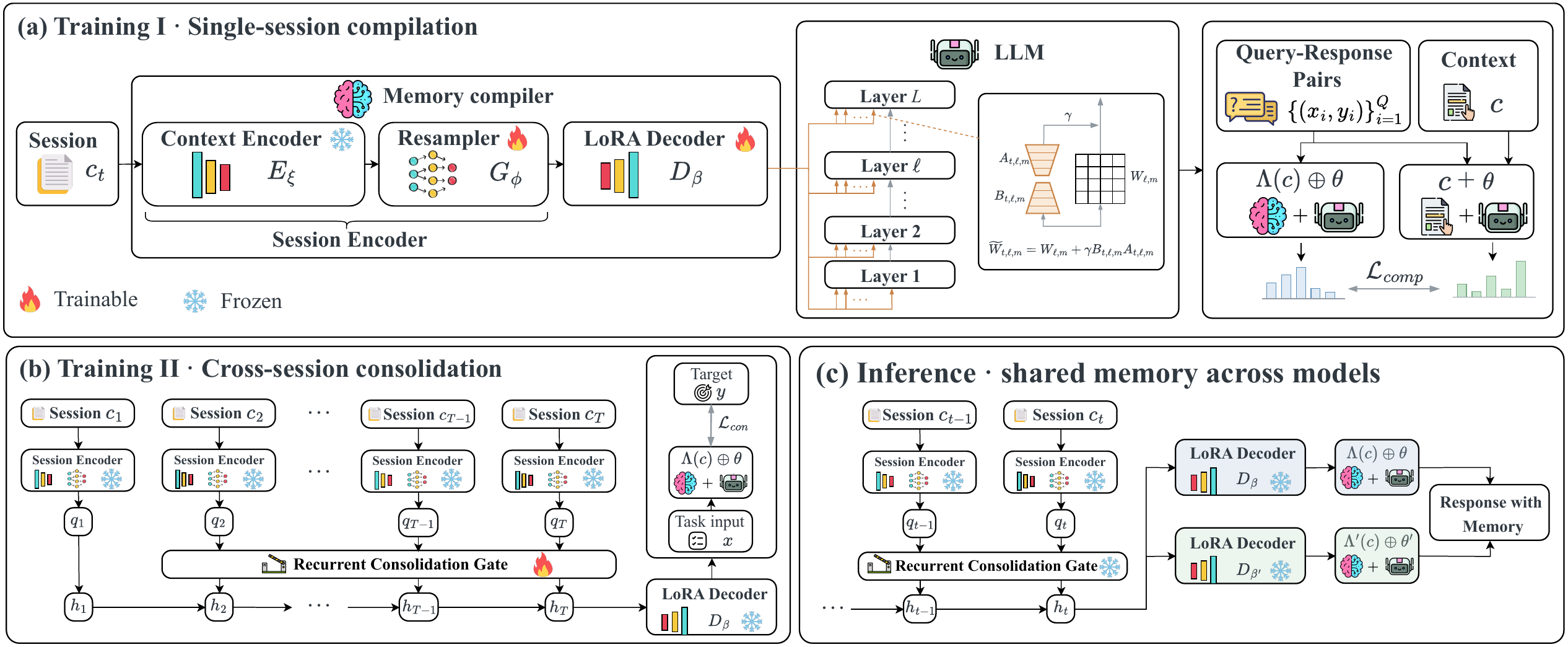}
    \caption{Overview of the \methodname{} architecture.
    (a) Single-session compilation trains the Perceiver-based resampler and
    LoRA decoder to match context-conditioned responses, with the context
    encoder and backbone frozen.
    (b) Cross-session consolidation trains only the recurrent gate to
    integrate session memories under task supervision.
    (c) At inference, accumulated memory is updated recurrently and decoded
    into backbone-specific LoRA parameters. Adapted decoders enable reuse of
    the same memory across backbones; all modules are frozen.}
    \label{fig:framework}
    \label{fig:two-phase-training}
\end{figure}

\begingroup
\setlength{\parskip}{3pt}
\begin{enumerate}[leftmargin=*,nosep,beginpenalty=10000,midpenalty=0]
    \item We formulate the problem of lifecycle-independent parametric memory and identify the requirements it imposes on memory writing, cross-session evolution, and backbone decoupling.
    \item We propose \methodname{}, to our knowledge the first lifecycle-independent parametric memory architecture, which maintains evolving cross-session memory in a model-independent latent space that is decoded into LoRA parameters for the current frozen backbone.
    \item Experiments across three benchmarks and five backbones, together with ablation and memory-dynamics analyses, validate the two-stage design and demonstrate that parametric memory can be decoupled from the backbone lifecycle.
\end{enumerate}
\endgroup

\section{Method}
\label{sec:method}

\subsection{Overview}
\label{sec:method-overview}

We consider an agent that receives an ordered stream of bounded interaction sessions $\mathcal{C}_{1:T}=(c_1,\ldots,c_T)$ and later answers a query $x$. Here, $T$ is the number of observed sessions, $t\in\{1,\ldots,T\}$ indexes their arrival order, and $c_t$ denotes the $t$-th session.  A frozen autoregressive language model $f_\theta$, whose backbone parameters are denoted by $\theta$, generates a response $y$ according to
\begin{equation}
    p(y\mid x,\mathcal{C}_{1:T})
    =\prod_{j=1}^{N}p(y_j\mid x,y_{<j},\mathcal{C}_{1:T}),
    \label{eq:task-formulation}
\end{equation}
where $N=|y|$ is the response length, $j\in\{1,\ldots,N\}$ is the generation position, $y_j$ is the token generated at position $j$, and $y_{<j}$ is its preceding response prefix. Our goal is a fixed-size persistent memory that carries information from $\mathcal{C}_{1:T}$ into generation and incorporates each new session without reprocessing earlier ones.

\methodname{} organizes memory processing into session encoding, recurrent consolidation, and model-specific decoding. \compilationphase{} (\autoref{fig:framework}(a)) learns to encode each session into fixed-shape latent memory and decode it into LoRA parameters that preserve the session's influence on the frozen backbone's responses. \consolidationphase{} (\autoref{fig:framework}(b)) then uses downstream supervision to learn how to selectively integrate successive session memories through a recurrent gate. The model-specific decoder maps the consolidated memory to LoRA parameters for response generation.

\subsection{\compilationphase}
\label{sec:memory-compiler}

The memory compiler is a hypernetwork comprising a session encoder and a model-specific LoRA decoder. \tbd{The session encoder combines a pretrained context encoder, frozen to maintain a stable feature space across backbone changes, with a trainable Perceiver-based resampler that compresses token features into fixed-shape session memory.} For a tokenized session $c_t$, these operations are
\begin{align}
    X_t &= E_\xi(c_t)\in\mathbb{R}^{L\times n_t\times d_e},
    \label{eq:context-features}\\
    q_t &= G_\phi(X_t)\in\mathbb{R}^{L\times M\times r\times d}.
    \label{eq:session-encoding}
\end{align}
Here, the frozen context encoder $E_\xi$, with parameters $\xi$ and hidden width $d_e$, extracts features for the $n_t$ tokens of $c_t$ from $L$ evenly spaced depths, one per adapted layer of the compilation backbone.  The axes of $X_t$ index encoder depth, token position, and feature.  We implement $G_\phi$, with trainable parameters $\phi$, as a Perceiver-based resampler~\citep{jaegle2021perceiver} that independently compresses each layer's token features using shared parameters.  In its output $q_t$, $M$ counts the types of linear modules receiving LoRA updates within each adapted layer.  The symbols $r$ and $d$ denote the LoRA rank and latent feature dimension, respectively.  Each slice $q_{t,\ell,m,k}\in\mathbb{R}^{d}$ represents rank component $k\in\{1,\ldots,r\}$ for module type $m\in\{1,\ldots,M\}$ at layer $\ell\in\{1,\ldots,L\}$.  The overall shape of $q_t$ is therefore independent of the session length $n_t$.

A decoder maps $q_t$ to a model-specific adapter collection and applies the resulting LoRA updates to the corresponding frozen backbone weights:
\begin{align}
    D_\beta(q_t)
    &=\Lambda_t=\{A_{t,\ell,m},B_{t,\ell,m}\}_{\ell,m},
    \label{eq:adapter-decoding}\\
    \widetilde W_{t,\ell,m}
    &=W_{\ell,m}+\gamma B_{t,\ell,m}A_{t,\ell,m}.
    \label{eq:generated-lora}
\end{align}
Here, $D_\beta$ denotes the decoder with trainable parameters $\beta$, and $\Lambda_t$ collects the generated factor pairs across layers $\ell$ and module types $m$. $A_{t,\ell,m}\in\mathbb{R}^{r\times d^{\mathrm{in}}_{\ell,m}}$ is the down-projection factor and $B_{t,\ell,m}\in\mathbb{R}^{d^{\mathrm{out}}_{\ell,m}\times r}$ is the up-projection factor generated from session $c_t$.  The symbols $d^{\mathrm{in}}_{\ell,m}$ and $d^{\mathrm{out}}_{\ell,m}$ denote the input and output widths of the original linear map $W_{\ell,m}\in\mathbb{R}^{d^{\mathrm{out}}_{\ell,m}\times d^{\mathrm{in}}_{\ell,m}}$.  The fixed coefficient $\gamma$ scales the LoRA update, and $\widetilde W_{t,\ell,m}$ denotes the resulting session-adapted map. We write $\theta\oplus\Lambda_t$ for the frozen backbone augmented with the generated adapter collection.

We train the memory compiler to encode session memories in latent space and make them available through generated LoRA parameters. For each session, we construct query--response pairs. Given the same query and reference response prefix, we align the next-token distribution of the frozen backbone equipped with the generated LoRA with that of the same backbone reading the session context without an adapter:
\begin{align}
    p^{\mathrm{ctx}}_{i,j}(v)
    &=p_\theta(v\mid c,x_i,y_{i,<j}),
    \label{eq:context-distribution}\\
    p^{\mathrm{mem}}_{i,j}(v)
    &=p_{\theta\oplus\Lambda(c)}(v\mid x_i,y_{i,<j}).
    \label{eq:compilation-distributions}
\end{align}
Here, each session $c$ has $Q$ query--response pairs $\{(x_i,y_i)\}_{i=1}^{Q}$, indexed by $i$, with a query $x_i$ and a fixed reference response $y_i=(y_{i,1},\ldots,y_{i,N_i})$ of length $N_i$. At response position $j\in\{1,\ldots,N_i\}$, $y_{i,j}$ denotes the token and $y_{i,<j}=(y_{i,1},\ldots,y_{i,j-1})$ its preceding prefix. The collection $\Lambda(c)=D_\beta(G_\phi(E_\xi(c)))$ is the adapter generated from session $c$.  The symbols $p^{\mathrm{ctx}}_{i,j}$ and $p^{\mathrm{mem}}_{i,j}$ denote the context-conditioned and adapter-conditioned next-token distributions, respectively, while $v$ indexes a candidate token in their shared output vocabulary.  Construction and split details are provided in \appref{app:training-data}.

We retain the $K$ highest-probability tokens under the context-conditioned distribution. Both distributions use this same token set and aggregate the remaining probability mass into a single tail category. The selected set, tail mass, and token-level forward-KL loss are
\begin{gather}
    S_{i,j}=\operatorname{TopK}_{v}\!\left(p^{\mathrm{ctx}}_{i,j}(v),K\right),
    \qquad \tau^a_{i,j}=1-\sum_{v\in S_{i,j}}p^a_{i,j}(v),
    \label{eq:tail-mass}\\
    \ell_{i,j}
    =\sum_{v\in S_{i,j}}p^{\mathrm{ctx}}_{i,j}(v)
      \log\frac{p^{\mathrm{ctx}}_{i,j}(v)}
               {p^{\mathrm{mem}}_{i,j}(v)}
      +\tau^{\mathrm{ctx}}_{i,j}
      \log\frac{\tau^{\mathrm{ctx}}_{i,j}}
               {\tau^{\mathrm{mem}}_{i,j}}.
    \label{eq:topk-tail-projection}
\end{gather}
Here, $\operatorname{TopK}_{v}(\cdot,K)$ returns the set of $K$ token indices with the highest probabilities, so $|S_{i,j}|=K$. The superscript $a\in\{\mathrm{ctx},\mathrm{mem}\}$ identifies one of the two distributions, $\tau^a_{i,j}$ is its probability mass outside $S_{i,j}$, and $\ell_{i,j}$ is the forward-KL loss at position $j$ of reference response $y_i$. The compiler averages $\ell_{i,j}$ over positions within each response and then over query--response pairs to obtain the session loss $\mathcal{L}_{\mathrm{FKL}}(c)$ and the regularized compilation objective $\mathcal{L}_{\mathrm{comp}}$:
\begin{align}
    \mathcal{L}_{\mathrm{FKL}}(c)
    &=\frac{1}{Q}\sum_{i=1}^{Q}\frac{1}{N_i}
      \sum_{j=1}^{N_i}\ell_{i,j},
    \label{eq:session-fkl}\\
    \mathcal{L}_{\mathrm{comp}}
    &=\mathbb{E}_{c}\!\left[
      \mathcal{L}_{\mathrm{FKL}}(c)+\lambda\mathcal{R}_1(c)
      \right].
    \label{eq:compilation-fkl}
\end{align}
Here, $\mathbb{E}_{c}$ averages over sessions in the compilation corpus, $\mathcal{R}_1(c)$ averages the absolute magnitudes of all LoRA factors decoded from $c$, and $\lambda$ is the regularization weight. Optimization updates $\phi$ and $\beta$ while holding $\xi$ and $\theta$ fixed.

The separation between $G_\phi$ and $D_\beta$ supports adaptation to a target backbone $f_{\theta'}$, whose frozen parameters are denoted by $\theta'$. We retain $G_\phi$, instantiate a target-specific decoder $D_{\beta'}$ with parameters $\beta'$, and optimize $\beta'$ with the same fixed-reference forward-KL objective while the memory encoder and target backbone remain fixed. The target decoder accepts the unchanged memory representation and generates LoRA factors matching the target backbone's layer count and module dimensions.

\subsection{\consolidationphase}
\label{sec:memory-consolidator}

Compilation places every $q_t$ in the same fixed-shape memory space.  Let $h_t\in\mathbb{R}^{L\times M\times r\times d}$ denote the accumulated memory after the first $t$ sessions.  We set $h_0=0$ before any session arrives and initialize the memory directly from the first session as $h_1=q_1$. For each subsequent session, a lightweight recurrent \consolidationgate{} combines the accumulated memory with the incoming representation through a retention tensor:
\begin{equation}
    z_t=\sigma\!\left([h_{t-1};q_t]W_g+b_g\right),\qquad t\geq2.
    \label{eq:consolidation-gate}
\end{equation}
The tensor $z_t\in\mathbb{R}^{L\times M\times r\times d}$ contains a $d$-dimensional retention vector at every layer, module, and rank index.  The operator $[\,;\,]$ denotes concatenation, $W_g\in\mathbb{R}^{2d\times d}$ is the shared gate projection, $b_g\in\mathbb{R}^{d}$ is its bias, and $\sigma$ is the element-wise sigmoid. Concatenation and the shared linear projection operate along the latent feature dimension. The gate updates the persistent state as
\begin{equation}
    h_t=z_t\odot h_{t-1}+(1-z_t)\odot q_t,\qquad t\geq2,
    \label{eq:consolidation-update}
\end{equation}
where $\odot$ denotes element-wise multiplication and $1$ is an all-ones tensor with the same shape as $z_t$.  Thus, $z_t$ weights the previous state and $1-z_t$ weights the incoming session representation. The same gate is shared across every layer, module, and rank index, yielding $2d^2+d$ parameters independent of $L$, $M$, and $r$.

Compilation learns to encode individual sessions as parametric memory through distribution matching. We separate consolidation learning from memory compilation to tailor memory retention and updating to the information needs and interaction patterns of the target application scenario. The second stage trains the gate with supervision from that scenario, and the learned policy is reused across subsequent interactions after deployment. For a training example $(\mathcal{C}_{1:T},x,y)$, the compiler produces the ordered representations $q_1,\ldots,q_T$, the recurrence forms the final persistent state $h_T$, and the decoder produces the corresponding adapter collection $\Lambda_T=D_\beta(h_T)$.  We collect the trainable gate parameters as $\psi=\{W_g,b_g\}$ and optimize the expected loss over the target application's training distribution:
\begin{equation}
    \mathcal{L}_{\mathrm{con}}(\psi)
    =\mathbb{E}_{(\mathcal{C}_{1:T},x,y)\sim\mathcal{D}_{\mathrm{task}}}
      \left[\ell_{\mathrm{task}}\!\left(
      p_{\theta\oplus\Lambda_T}(\,\cdot\mid x),y\right)\right],
    \label{eq:phase2-objective}
\end{equation}
where $\mathcal{D}_{\mathrm{task}}$ is the training-data distribution of the target application, $p_{\theta\oplus\Lambda_T}(\,\cdot\mid x)$ is the memory-conditioned response distribution, and $\ell_{\mathrm{task}}$ is a differentiable prediction loss that evaluates this prediction against the target $y$. Gradients propagate through the frozen backbone and decoder to update $\psi$, while $\theta$, $\xi$, $\phi$, and $\beta$ remain fixed.  The frozen compiler also allows the $q_t$ representations to be precomputed during training. At deployment (\autoref{fig:framework}(c)), all modules are frozen. The first session initializes memory, and subsequent sessions update it through encoding and recurrent consolidation.  Queries access the retained memory through the current backbone's decoder.  Backbone replacement preserves the accumulated memory, session encoder, and learned gate, switching only to the new backbone and its adapted decoder. Subsequent sessions use the same memory space and consolidation policy.

\section{Experiments}
\label{sec:experiments}

We evaluate three aspects of \methodname{}: cross-session memory performance and generalization, the contributions of its two training stages, and the lifecycle properties required for long-term deployment.

\subsection{Experimental Setup}
\label{sec:experimental-setup}

\noindent\textbf{Benchmarks and metrics.}\quad \permabench~\citep{liu2026perma} evaluates memory formation, revision, and intervention across ordered sessions; we report its four clean and noisy single-domain (SD) and multi-domain (MD) settings and their macro-average. \personamembench~\citep{jiang2025personamemv2} derives independent queries from long user histories; we report Overall, Self, and Current accuracy. \prefevalbench~\citep{zhao2025prefeval} measures memory after 10, 70, and 300 intervening turns, averaged across its three information forms.  Detailed task construction and metric definitions appear in \appref{app:benchmark-protocols}.

\noindent\textbf{Baselines.}\quad Textual-memory methods include Full Context, RAG~\citep{lewis2020rag} with M3-Embedding~\citep{chen2024m3embedding}, Rolling Summary, Mem0~\citep{chhikara2025mem0}, and LightMem~\citep{fang2026lightmem}. Parametric baselines include SFT, trained on the same downstream split with the visible history, and the released Metis-9B recurrent-memory model~\citep{zhang2026metis}.  No Context and Gold State provide query-only and annotated-memory references.  Implementations and method-specific hyperparameters are documented in \appref{app:baseline-implementations}.

\noindent\textbf{Protocol.}\quad In the main comparison, \methodname{} and our baseline implementations use Qwen3-8B~\citep{yang2025qwen3} with thinking disabled. We directly evaluate the released Metis-9B model. Generalization and transfer span five backbones across model families, scales, and architectures. \permabench{} uses ten-fold leave-one-user-out evaluation, while \personamembench{} and \prefevalbench{} follow their official splits. For these benchmarks, we train the consolidation gate on each training split with the compiler from \autoref{sec:memory-compiler} and answer-label cross-entropy. Full training and inference protocols appear in \appref{app:evaluation-protocol}.

\subsection{Main Results}
\label{sec:main-results}

\autoref{tab:main-results} compares all methods on cross-session memory benchmarks. Bold and underlined values mark the best and second-best non-oracle results, excluding Gold State.

\begin{table}[htbp]
    \centering
    \caption{Main comparison on cross-session memory benchmarks (accuracy, \%).}
    \label{tab:main-results}
    \vspace{2pt}
    {\scriptsize
    \renewcommand{\arraystretch}{0.94}
    \setlength{\tabcolsep}{1.7pt}
    \resizebox{0.98\linewidth}{!}{%
    \begin{tabular}{ll*{11}{r}}
        \toprule
        \multirow{2}{*}{\textbf{Type}} & \multirow{2}{*}{\textbf{Method}} &
        \multicolumn{5}{c}{\textbf{\permabench}} &
        \multicolumn{3}{c}{\textbf{\personamembench}} &
        \multicolumn{3}{c}{\textbf{\prefevalbench}} \\
        \cmidrule(lr){3-7}\cmidrule(lr){8-10}\cmidrule(lr){11-13}
        & & \textbf{SD-C} & \textbf{SD-N} & \textbf{MD-C} & \textbf{MD-N} &
        \textbf{Core} & \textbf{Overall} & \textbf{Self} & \textbf{Current} &
        \textbf{10} & \textbf{70} & \textbf{300} \\
        \midrule
        \multirow{2}{*}{Reference}
        & No Context & 66.73 & 66.73 & 54.60 & 54.60 & 60.66 & 25.74 & 25.28 & 25.42 & 36.67 & 39.82 & 39.07 \\
        & Gold State & 72.70 & 72.23 & 66.06 & 65.47 & 69.12 & 57.90 & 64.05 & 69.16 & 80.18 & 80.74 & 80.00 \\
        \midrule
        \multirow{5}{*}{Textual}
        & Rolling Summary & 39.57 & 40.27 & 23.34 & 27.41 & 32.65 & 26.58 & 25.46 & 25.32 & 44.26 & 45.19 & 45.19 \\
        & LightMem & 56.63 & 56.20 & 40.72 & 40.20 & 48.44 & 30.46 & 31.42 &
        34.51 & \textbf{81.30} & \textbf{79.07} & \underline{73.52} \\
        & Mem0 & 57.57 & 58.57 & 41.03 & 42.10 & 49.82 & 30.02 & 30.26 & 31.90 & \underline{73.15} & 65.19 & 61.67 \\
        & RAG (BGE-M3) & 63.00 & 62.53 & 49.16 & 47.46 & 55.54 & 30.32 & 31.13 & 31.70 & 59.26 & 54.63 & 50.74 \\
        & Full Context & 76.47 & 73.03 & 70.72 & 69.95 & 72.54 & 31.10 &
        32.72 & 33.97 & 60.00 & 45.00 & 40.56 \\
        \midrule
        \multirow{3}{*}{Parametric}
        & SFT & 78.00 & 71.47 & 49.30 & 42.61 & 60.35 &
        \underline{31.64} & \underline{33.65} & \underline{35.47} & 57.22 & 40.00 & 38.15 \\
        & Metis-9B & \underline{80.87} & \underline{80.47} &
        \textbf{80.17} & \underline{79.29} & \underline{80.20} & 29.64 & 30.24 & 31.32 & 43.89 & 39.07 & 39.82 \\
        \rowcolor{black!6}
        \cellcolor{white} & \textbf{\methodname{} (ours)} & \textbf{90.57} & \textbf{91.20} &
        \underline{79.67} & \textbf{80.62} & \textbf{85.52} &
        \textbf{40.22} & \textbf{43.61} & \textbf{47.28} & 69.81 &
        \underline{72.41} & \textbf{74.81} \\
        \bottomrule
    \end{tabular}}}
\end{table}

Across \permabench{} and \personamembench{}, \methodname{} obtains the strongest non-oracle aggregate performance under two independently constructed cross-session evaluations. On \permabench{}, it achieves \permacoreacc{}, outperforming Metis-9B and Full Context by 5.32 and 12.98 percentage points (pp), respectively, and ranks first in three of the four core settings. The advantage over Metis also holds with Qwen3.5-9B for both methods (\appref{app:same-backbone-comparison}). \methodname{} also surpasses SFT by 25.17 pp under the same downstream split and visible-history boundary, extending its advantage to a parametric baseline trained on in-domain histories. \methodname{} also achieves the highest average across all seven historical-context variants at \permasevenacc{} (\appref{app:perma-style-results}). On \personamembench{}, it achieves \personamemacc{} Overall accuracy, outperforming SFT by 8.58 pp and Full Context by 10.89 and 13.31 pp on Self and Current. The consistent advantage across benchmark constructions shows that the learned memory remains effective while tracking currently valid user information over long histories.

\prefevalbench{} further reveals how the methods respond as the intervening history grows.  \methodname{} ranks third at 10 turns and second at 70 turns, then achieves the best 300-turn result at 74.81\%, surpassing LightMem by 1.29 pp. Over the same range, LightMem declines from 81.30\% to 73.52\%, whereas \methodname{} rises from 69.81\% to 74.81\%.  Its strongest relative performance at the longest interval shows that consolidated memory remains effective when the preference-bearing interaction is separated from the query by a long history.

\begin{figure}[htbp]
    \centering
    \includegraphics[width=0.98\linewidth]{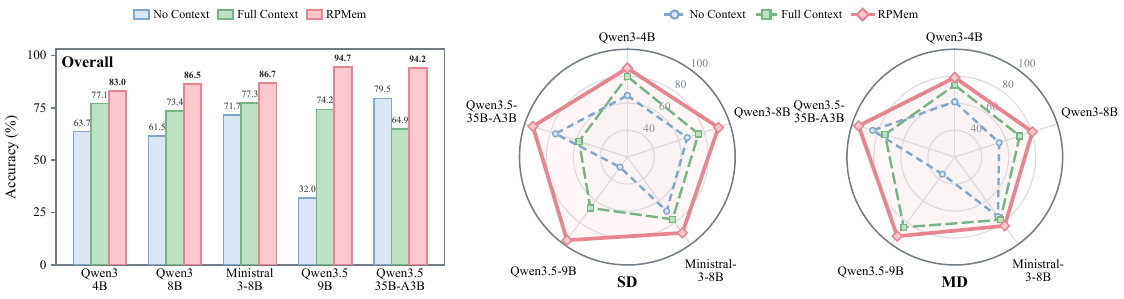}
    \caption{\permabench{} generalization across five backbones.  Left:
    seven-variant average; right: SD and MD averages (accuracy, \%).}
    \label{fig:model-results}
\end{figure}

\noindent\textbf{Performance across language models.}\quad We evaluate \methodname{} on five backbones against matched No Context and Full Context references. \autoref{fig:model-results} shows consistent gains across model families, sizes, and dense and MoE architectures, with overall improvements of 5.95--20.41 pp over the stronger reference for each backbone. Full numerical results are reported in \appref{app:model-generalization-results}. These findings support the generalizability of the memory mechanism across heterogeneous language-model backbones.

\FloatBarrier

\subsection{Ablation Studies}
\label{sec:ablations}

\methodname{} learns memory at two levels: the compiler encodes each session, and the recurrent consolidation module integrates the resulting memories over time.  We isolate these components by varying the compiler objective under a fixed consolidation protocol and then varying the consolidation rule with the selected compiler.  \autoref{tab:ablation-summary} reports both ablations with SD and MD macro-averages; complete definitions and setting-level results appear in \appref{app:ablation-details}.

\begin{table}[htbp]
    \centering
    \begin{minipage}{0.72\linewidth}
    \centering
    \caption{Ablations on \permabench{} (accuracy, \%):
    \textbf{(a)} compilation objectives and \textbf{(b)} consolidation rules.
    Core SD and Core MD average the clean and noisy settings.
    FKL/RKL denote forward/reverse KL; bold and underline mark the best and
    second-best results in each panel.}
    \label{tab:ablation-summary}
    \small
    \renewcommand{\arraystretch}{1.0}
    \setlength{\tabcolsep}{4pt}
    \begin{tabular*}{\linewidth}{@{\extracolsep{\fill}}c|ccc@{}}
        \toprule
        Variant & Core SD & Core MD & Core Avg. \\
        \midrule
        \multicolumn{4}{l}{\textit{(a) Memory compilation objective}} \\
        \compilersft{} & 72.20 & 50.23 & 61.21 \\
        \sampledrkl{} (top-$K$) & 74.08 & 59.73 & 66.91 \\
        \sampledrkl{} (K3) & 70.82 & 63.22 & 67.02 \\
        Top-$K$ CE & 82.99 & 72.34 & 77.66 \\
        \sampledfkl{} & \underline{88.37} & \underline{79.14} &
        \underline{83.75} \\
        \textbf{\fixedfkl{}} & \textbf{90.89} & \textbf{80.15} &
        \textbf{85.52} \\
        \midrule
        \multicolumn{4}{l}{\textit{(b) Cross-session consolidation}} \\
        Rank concatenation & 14.97 & 12.71 & 13.84 \\
        Latent averaging & 62.40 & 28.64 & 45.52 \\
        Latest Session & 67.72 & \underline{39.40} & 53.56 \\
        Factor averaging & \underline{69.44} & 38.99 & \underline{54.21} \\
        \textbf{Learned gate (ours)} & \textbf{90.89} & \textbf{80.15} &
        \textbf{85.52} \\
        \bottomrule
    \end{tabular*}
    \end{minipage}
\end{table}

\noindent\textbf{Compilation objective.}\quad Our compilation objective matches the context-conditioned reference distribution along fixed reference-response prefixes using forward KL (\fixedfkl{}). It reaches 85.52\%, improving over selected-token cross-entropy (Top-$K$ CE) by 7.86 pp and hard-label supervision (\compilersft{}) by 24.31 pp. The 16.45 pp gain from hard labels to selected-token probabilities shows that the relative likelihoods of plausible continuations carry information induced by the session.  Retaining the probability mass outside the selected support contributes a further 7.86 pp, showing that faithful compilation also depends on distributional coverage.  \sampledfkl{} remains within 1.77 pp of \fixedfkl{}, whereas reversing the KL direction reduces accuracy by more than 16 pp under sampled-response training.  Together, these results favor matching the full context-conditioned distribution along a fixed, grounded response trajectory.  Objective definitions and checkpoint-wise training dynamics are provided in \appref{app:compilation-objectives} and \appref{app:compiler-dynamics}.

\noindent\textbf{Cross-session consolidation.}\quad The learned gate reaches 85.52\%, exceeding the strongest evaluated fixed rule, LoRA factor averaging, by 31.31 pp.  Factor averaging improves over retaining only the latest session by 0.65 pp, latent averaging performs 8.04 pp worse, and rank concatenation falls to 13.84\% despite retaining every decoded session adapter. The learned gate gains 21.45 pp over factor averaging in SD histories and 41.16 pp in MD histories, localizing its largest benefit to sequences containing heterogeneous memory.  These comparisons show the benefit of task-supervised, coordinate-wise integration over the evaluated fixed rules for combining accumulated and incoming information.  Further definitions and data-efficiency results appear in \appref{app:consolidation-variants} and \appref{app:consolidation-supervision}. Additional experiments on Clean SD and Clean MD show consistent gains over update averaging, tuned EMA, and a task-trained scalar gate (\appref{app:additional-consolidation}).

\tbd{A zero-shot cross-benchmark evaluation further tests the learned consolidation policy: PERMA-trained gates reach 38.14\% on PersonaMem-v2, compared with 35.25\% for latent averaging and 45.73\% for a target-trained gate (Appendix~\ref{sec:additional-gate-transfer}).}

\subsection{Lifecycle Properties}
\label{sec:lifecycle-properties}

We first test the defining lifecycle claim that accumulated memory remains reusable after backbone replacement.  We then evaluate its deployment viability through resource scaling and general-capability retention.

\noindent\textbf{Cross-backbone reuse.}\quad We reuse the frozen Qwen3-8B session encoder with Qwen3-4B, Ministral-3-8B~\citep{mistralai2026ministral3}, Qwen3.5-9B, and Qwen3.5-35B-A3B~\citep{qwen2026qwen35}, adapting only the model-specific decoder during compilation for each target. Under the same target-side training budget, transfer improves all 28 backbone--setting combinations over compilers trained from scratch and raises the cross-model average from 75.97\% to 89.64\%, showing that one parametric memory representation can serve different backbones through target-specific decoding. Complete results and adaptation costs appear in \appref{app:backbone-transfer}.

\tbd{We additionally tested continued memory updates after a mid-history backbone switch while retaining the source-trained consolidation gate and accumulated memory. On PERMA Noisy SD, accuracy was 85.37\% for Qwen3-4B to Qwen3-8B and 76.02\% for the reverse direction, with only the serving backbone and decoder replaced (Appendix~\ref{sec:additional-backbone-continuation}).}

\noindent\textbf{Streaming deployment.}\quad We measure update latency, persistent-memory size, and historical query tokens as the session count increases from 1 to 64, comparing \methodname{} with Full Context, Mem0, and uncompressed adapter rank concatenation. \autoref{tab:deployment-cost} reports values at 64 sessions and growth over this range. Metric definitions and the full protocol are provided in \appref{app:efficiency-details}.

\begin{table}[tb]
    \centering
    \caption{Online deployment efficiency at 64 accumulated sessions.}
    \label{tab:deployment-cost}
    \footnotesize
    \renewcommand{\arraystretch}{0.96}
    \setlength{\tabcolsep}{2pt}
    \begin{tabular*}{\linewidth}{@{\extracolsep{\fill}}ccccccc@{}}
        \toprule
        \multirow{2}{*}{Type} & \multirow{2}{*}{Method} &
        \multicolumn{2}{c}{Update} & \multicolumn{2}{c}{Memory} &
        \multirow{2}{*}{\shortstack{Hist. tokens\\$\downarrow$}} \\
        \cmidrule(lr){3-4}\cmidrule(lr){5-6}
        & & Time (s) $\downarrow$ & Growth $\downarrow$ & Size (MiB) $\downarrow$ &
        Growth $\downarrow$ & \\
        \midrule
        \multirow{2}{*}{Textual}
        & Full Context & -- & -- & \textbf{0.138} & $78.09\times$ & 28,498.4 \\
        & Mem0 & 12.554 & $1.10\times$ & 20.041 & $62.11\times$ & 441.7 \\
        \midrule
        \multirow{2}{*}{Parametric}
        & Adapter rank concatenation & 0.948 & $29.46\times$ & 183.738 &
        $5.10\times$ & \textbf{0} \\
        & \textbf{\methodname{} (ours)} & \textbf{0.043} &
        \textbf{$1.03\times$} & 36.562 & \textbf{$1.00\times$} & \textbf{0} \\
        \bottomrule
    \end{tabular*}
\end{table}

At 64 sessions, \methodname{} writes a session in 0.043 seconds, approximately 22 times faster than rank concatenation and 292 times faster than Mem0.  Across the full profile, update and memory growth remain $1.03\times$ and $1.00\times$, respectively, with no historical query tokens.  Avoiding history reprocessing also reduces query cost: over the 5,000-question \personamembench{} test set with approximately 32K-token histories, mean language-model forward time is 0.055 seconds, compared with 4.287 seconds for Full Context. Training the shared compiler requires an estimated 338 GPU-hours, whereas consolidation for a representative \permabench{} fold takes 13 minutes on a single GPU; complete costs appear in \appref{app:efficiency-details}.

\noindent\textbf{General-capability retention.}\quad Finally, we evaluate the unmodified backbone (Base LLM), Latest Session, and \methodname{} on MMLU~\citep{hendrycks2021mmlu}, GSM8K~\citep{cobbe2021gsm8k}, and IFEval~\citep{zhou2023ifeval}, averaging the two memory methods over ten user-specific adapters.  Complete protocols and user-level results appear in \appref{app:efficiency-details}.

\begin{table}[htbp]
    \centering
    \begin{minipage}{0.72\linewidth}
    \centering
    \caption{General-capability retention (accuracy, \%).}
    \label{tab:capability-retention}
    \small
    \renewcommand{\arraystretch}{0.98}
    \setlength{\tabcolsep}{3pt}
    \begin{tabular*}{\linewidth}{@{\extracolsep{\fill}}cccc@{}}
        \toprule
        Method & MMLU & GSM8K & IFEval \\
        \midrule
        Base LLM & 71.96 & 75.21 & 81.70 \\
        Latest Session & 69.85 & 73.15 & 77.38 \\
        \methodname{} (ours) & 70.13 & 73.78 & 77.89 \\
        \bottomrule
    \end{tabular*}
    \end{minipage}
\end{table}

\methodname{} remains within 1.83 pp of Base LLM on MMLU, 1.43 pp on GSM8K, and 3.81 pp on IFEval, while improving over Latest Session on all three benchmarks.  Cross-session memory therefore largely preserves the backbone's general capabilities.

\section{Analysis of Learned Consolidation Dynamics}
\label{sec:memory-analysis}

The experiments demonstrate the effectiveness of \methodname{}'s cross-session memory. We then examine how the learned consolidation mechanism organizes memory as sessions accumulate. We study the Qwen3-8B configuration of \methodname{} under the \permabench{} ten-fold protocol.  \tbd{For each held-out user in Clean SD and Noisy SD, we replay the query-visible history of each task-conditioned trajectory through the recurrent gate to track how sessions modify and retain memory. Because trajectories may share historical sessions, statistics are first aggregated within each held-out user and then averaged across the ten users (\appref{app:dynamics-protocol}).}

\subsection{Semantically Structured Memory Updates}
\label{sec:structured-updates}

\tbd{To understand how the recurrent gate manages memory across sessions, we examine how strongly different events write to the accumulated memory.} \tbd{For each subsequent session}, the gate determines how much of the existing memory to retain and how much weight to assign to the incoming session representation. \tbd{We quantify this balance using write magnitude, defined as the average weight assigned to the incoming memory across all latent coordinates. For events spanning multiple compiler segments, we combine their successive updates before computing this measure. We compare two event types annotated in \permabench{}: domain emergence, which introduces a new preference domain, and supplement, which adds information to an existing domain.} Emergence produces larger writes for every held-out user, and the emergence--supplement write difference is approximately five times larger than the difference between matched Clean and Noisy inputs (\autoref{fig:memory-dynamics}(a--b)).  Write magnitude therefore follows an event's contribution to the memory timeline while remaining stable under surface perturbations.

\tbd{To examine whether events in the same domain update similar memory coordinates, we compare their write patterns---the incoming-memory weights across latent coordinates---using cosine similarity after centering. We also measure one-step source survival, the fraction of a historical session's contribution retained after an incoming event. Same- and cross-domain comparisons use pairs matched within each user's trajectories by event timing and length (Appendix~F.4).} Same-domain events have more similar write patterns (\autoref{fig:memory-dynamics}(c)), while a new event attenuates same-domain historical contributions more strongly than cross-domain contributions (\autoref{fig:memory-dynamics}(d)). Both paired differences hold for every held-out user (\autoref{fig:memory-dynamics}(e)), indicating that learned consolidation combines domain-specific revision with cross-domain preservation.

\begin{figure}[tbp]
    \centering
    \includegraphics[width=\linewidth]{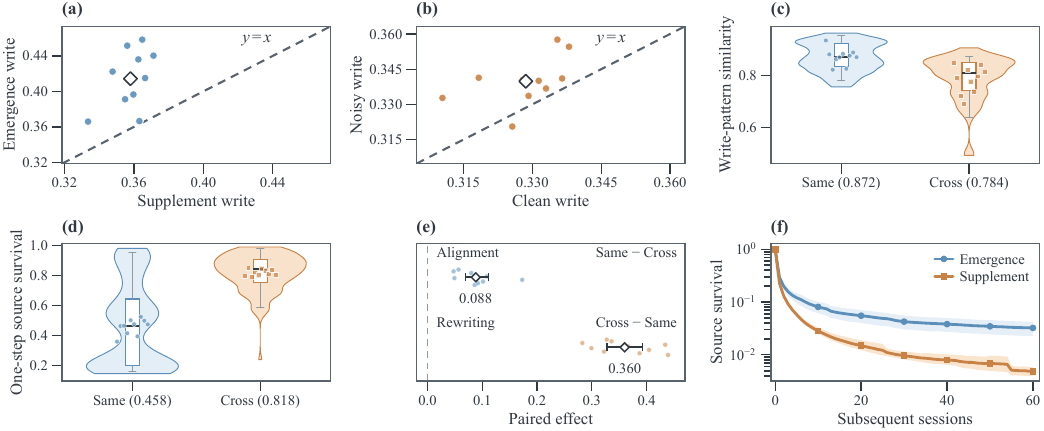}
    \caption{Learned consolidation dynamics on \permabench{}.
    (a--b) User-mean write magnitudes by event role and noise condition;
    diamonds mark across-user means, and dashed diagonals indicate equality.
    (c--d) Same- versus cross-domain write-pattern similarity and one-step
    source survival; violins show matched-pair distributions and points show
    user means. (e) Paired differences: Same minus Cross for similarity,
    Cross minus Same for survival. (f) Source survival over subsequent updates
    on a logarithmic scale. Statistical details, including interval definitions,
    appear in \appref{app:memory-dynamics-details}.}
    \label{fig:memory-dynamics}
\end{figure}

\subsection{Long-Term Survival of Memory Sources}
\label{sec:source-survival}

We next examine whether these differentiated updates preserve memory over longer horizons. \tbd{We track the fraction of each session's initial contribution that remains after subsequent updates, extending the one-step measure to longer horizons.} The survival advantage of domain-emergence events grows from $1.31\times$ after one update to $6.64\times$ after 60 (\autoref{fig:memory-dynamics}(f)).  \tbd{To assess functional retention separately from coefficient-based source survival, we compare answer accuracy immediately after the target memory event (T2) and after subsequent intervening sessions (T3). Across seven \permabench{} variants, accuracy changes from 86.7\% at T2 to 86.5\% at T3 (\appref{app:memory-dynamics-details}).}  Across the observed histories, fixed-size memory exhibits differentiated writing and retention, with domain-establishing events receiving stronger writes and persisting longer than supplements.  A representative held-out trajectory in \appref{app:gate-cases} traces this organization through a complete interaction history.

\section{Related Work}
\label{sec:related-work}

\noindent\textbf{Text-Based Agent Memory.}
Most agent-memory systems maintain interaction history as external textual
records and retrieve selected entries for each query.  Generative
Agents~\citep{park2023generative} organizes experience as a memory stream,
MemGPT~\citep{packer2023memgpt} manages hierarchical storage, and Mem0 and
LightMem~\citep{chhikara2025mem0,fang2026lightmem} extract structured records
from conversations.  A-MEM~\citep{xu2025amem} constructs linked notes, while
Infini Memory~\citep{ji2026infinimemory} consolidates observations into topic
documents.  Memory-R1 and AgeMem~\citep{yan2026memoryr1,yu2026agemem} learn
memory management through reinforcement learning, with AgeMem jointly
controlling external memory and the active context.
GRU-Mem~\citep{sheng2026grumem} learns when to update textual memory and stop
reading further context.
At the infrastructure level, MemOS~\citep{li2025memos} unifies the
organization and lifecycle management of textual, activation, and parametric
memories; \methodname{} addresses the learned representation and integration
of cross-session parametric memory.
These advances improve storage and retrieval, yet
cross-session knowledge must still be selected and reconstructed within a
finite context on every call, making long-horizon performance jointly dependent
on retrieval quality, compression fidelity, and contextual reasoning.

\noindent\textbf{Parametric Agent Memory.}
Parametric memory makes prior information available through model computation
rather than context injection.  PAM~\citep{tang2026pam} internalizes
long-horizon agent experience into LoRA adapters through fine-tuning;
TMEM~\citep{ren2026tmem} accumulates online LoRA gradient updates from extracted
experience.
SELF-PARAM~\citep{wang2025selfparam} integrates successive contexts into model
weights by matching context-conditioned response distributions.
WISE and ELDER~\citep{wang2024wise,li2025elder} manage sequential knowledge
edits through routed side memory and mixtures of LoRA adapters, respectively.
EVAF~\citep{han2026memorydepth} selects events using surprise and valence
signals, then updates a LoRA adapter with buffered replay. These approaches
use gradient-based writes; \methodname{} learns compilation and consolidation
offline and performs subsequent memory updates through forward computation.
ParamMem~\citep{yao2026parammem} encodes cross-sample reflection patterns into
model parameters.  MLP Memory~\citep{wei2025mlpmemory} distills a $k$NN
retriever's output distribution and combines its predictions with the language
model's.  \mbox{Titans~\citep{behrouz2025titans}} updates a neural memory module
at test time, while GradMem~\citep{kuratov2026gradmem} optimizes prefix memory
tokens using a context-reconstruction objective with the language model frozen.
Architecture-native methods maintain mutable memory inside
the language model: MEMORYLLM~\citep{wang2024memoryllm} introduces a
self-updatable latent memory pool, M+~\citep{wang2025mplus} extends it with
scalable long-term retrieval, and Metis~\citep{zhang2026metis} learns
forward-updated memory matrices through memory-specific mid-training.  Metis
represents the closest concurrent work: it satisfies both forward-only writing
and cross-session evolution through gated memory dynamics, yet its memory
dimensions, update projections, and read-write mechanisms are defined by the
specific backbone architecture, preventing memory reuse after model
replacement.  These architecture-native approaches couple memory to the
backbone that produced it.

\noindent\textbf{Context-to-Adapter Generation and Transfer.}
Context distillation optimizes parameters to reproduce behavior induced by a
given context~\citep{snell2022contextdistillation}.
MAC~\citep{tack2024mac} amortizes documents into a memory bank and aggregates
their modulations for each query. \methodname{} consolidates incoming
sessions into persistent, fixed-size memory before a query arrives.
Doc-to-LoRA~\citep{charakorn2026doctolora} amortizes document-specific
distillation into a shared Perceiver hypernetwork that generates LoRA
parameters in one forward pass; SHINE~\citep{liu2026shine} scales this mapping
to multi-layer adapters, and Profile-to-PEFT~\citep{yang2026profilepeft}
generates personalized adapters from textual user profiles.  These methods
establish forward mappings from bounded text to model parameters, but each
updated source produces a new static adapter without a mechanism for
cross-session memory evolution.  Generative
Adapter~\citep{chen2025generativeadapter} supports forward-only adaptation to
streaming context through backbone-specific weight updates, and
LoRA-Gen~\citep{xiao2025loragen} generates adapters for a target model from task
descriptions.  Trans-LoRA, Trans-PEFT, and
TiTok~\citep{wang2024translora,gu2025transpeft,jung2026titok} transfer adapters
learned from fixed task data across backbones, addressing model portability for
static adapters rather than evolving memory.
Memory Decoder~\citep{cao2025memorydecoder} learns a retriever-derived
prediction distribution that can be combined with backbones sharing a
tokenizer, enabling cross-model reuse of domain memory. \methodname{}
complements this portability with recurrent cross-session integration and
model-specific decoding of accumulated memory into LoRA parameters.

\section{Conclusion}
\label{sec:discussion}
\label{sec:conclusion}

We introduced \methodname{}, a recurrent parametric memory architecture that decouples accumulated experience from the serving backbone. By combining session compilation with learned consolidation, it maintains fixed-size cross-session memory without adding historical text to the query context. Experiments on three memory benchmarks demonstrate strong performance, while ablations and memory-dynamics analyses support the role of selective integration within a fixed capacity. Evaluations across five backbones demonstrate broad applicability, and backbone-replacement experiments show that accumulated memory can be reused through model-specific decoding. Deployment measurements further indicate approximately constant update cost and per-user memory size as histories grow, with limited degradation in general capabilities. Together, these results support lifecycle-independent parametric memory that accumulates experience across sessions and remains usable as serving backbones evolve.

\section{Limitations}
\label{sec:limitations}

Application-specific consolidation aligns memory retention and updating with the target scenario's requirements, but the learned policy's applicability to scenarios with substantially different memory needs remains to be studied. Future work could explore cross-scenario learning that balances shared consolidation capabilities with application-specific adaptation.

\section*{Acknowledgments}
This work was supported by Qwen Business Unit through Alibaba Research Intern Program.

\par\endgroup

\bibliography{references}
\bibliographystyle{unsrtnat}

\clearpage
\raggedbottom
\renewcommand{\floatpagefraction}{0.85}
\setcounter{topnumber}{4}
\setcounter{totalnumber}{6}
\makeatletter
\setlength{\@fptop}{0pt}
\setlength{\@fpsep}{12pt plus 2pt minus 2pt}
\setlength{\@fpbot}{0pt plus 1fil}
\makeatother
\appendix
\section*{Appendix}
\numberwithin{equation}{section}
\numberwithin{figure}{section}
\numberwithin{table}{section}

\FloatBarrier
\section{Algorithms}
\label{app:algorithms}
\numberwithin{algorithm}{section}
\providecommand{\algorithmautorefname}{Algorithm}
\begingroup
\raggedbottom

The following algorithms specify the two training stages and the online
memory interface using the notation of \autoref{sec:method}.
The context encoder $E_\xi$ and answering backbone $f_\theta$ are frozen
throughout. Compilation optimizes the Perceiver-based resampler parameters $\phi$ and
decoder parameters $\beta$; consolidation subsequently optimizes the gate
parameters $\psi=\{W_g,b_g\}$.  The training data are constructed in
\appref{app:training-data}, and numerical configurations are given in
\appref{app:compiler-configuration}.
We write $\operatorname{AdamW}(w,g;s)$ for an optimizer step with parameters
$w$, gradient $g$, and optimizer state $s$, returning the updated parameters
and state.  The state contains the moment estimates and step count and is
initialized to zero.  All optimizer hyperparameters follow the configuration
appendix.

\noindent\textbf{Session compilation.}\quad
Let $\mathcal{D}_{\mathrm{comp}}$ denote the collection of training sessions
and their fixed query--response pairs.  For each pair, the reference model
receives the session as text and is evaluated on the fixed response prefix.
In \autoref{alg:compilation}, $\operatorname{TopK}_v(p,K)$ returns the indices
of the $K$ largest probabilities in distribution $p$.
The projection $\Pi_S(p)=((p(v))_{v\in S},\,1-\sum_{v\in S}p(v))$ collects
the probabilities on support $S$ and their complementary tail mass.
The cache $\mathcal{T}(c,i,j)$ associates session $c$, pair $i$, and response
position $j$ with this support and projected reference distribution; its
probabilities are recovered from the stored reference log-probabilities.
We denote the cached projected distribution by $\bar p^{\mathrm{ctx}}_{i,j}$.
The symbols $\mathcal{B}$, $\mathcal{L}_c$, and
$\widehat{\mathcal{L}}_{\mathrm{comp}}$ denote a mini-batch, its individual
session objectives, and their mean, respectively.  The number of compilation
epochs is denoted by $E_{\mathrm{comp}}$; the reference cache is prepared once
and reused across these epochs.  Factor regularization
$\mathcal{R}_1$ follows \autoref{eq:factor-regularization}.

\begin{algorithm}[htbp]
    \caption{Single-session memory compilation}
    \label{alg:compilation}
    \small
    \begin{algorithmic}[1]
        \Require $\mathcal{D}_{\mathrm{comp}}, E_\xi,f_\theta,G_\phi,D_\beta,K,\lambda,E_{\mathrm{comp}}$
        \Ensure $\phi,\beta$
        \State $s_{\mathrm{comp}}\gets0$
        \For{$(c,\{(x_i,y_i)\}_{i=1}^{Q})\in\mathcal{D}_{\mathrm{comp}}$}
            \For{$i=1,\ldots,Q;\ j=1,\ldots,N_i$}
                \State $p^{\mathrm{ctx}}_{i,j}\gets p_\theta(\cdot\mid c,x_i,y_{i,<j})$
                \State $S_{i,j}\gets\operatorname{TopK}_v(p^{\mathrm{ctx}}_{i,j},K)$
                \State $\mathcal{T}(c,i,j)\gets(S_{i,j},\Pi_{S_{i,j}}(p^{\mathrm{ctx}}_{i,j}))$
            \EndFor
        \EndFor
        \For{$e=1,\ldots,E_{\mathrm{comp}}$}
            \For{each mini-batch $\mathcal{B}\subset\mathcal{D}_{\mathrm{comp}}$}
                \For{$(c,\{(x_i,y_i)\}_{i=1}^{Q})\in\mathcal{B}$}
                    \State $q\gets G_\phi(E_\xi(c))$
                    \State $\Lambda(c)\gets D_\beta(q)$
                    \For{$i=1,\ldots,Q;\ j=1,\ldots,N_i$}
                        \State $(S_{i,j},\bar p^{\mathrm{ctx}}_{i,j})\gets\mathcal{T}(c,i,j)$
                        \State $p^{\mathrm{mem}}_{i,j}\gets p_{\theta\oplus\Lambda(c)}(\cdot\mid x_i,y_{i,<j})$
                        \State $\ell_{i,j}\gets D_{\mathrm{KL}}(\bar p^{\mathrm{ctx}}_{i,j}\,\|\,\Pi_{S_{i,j}}(p^{\mathrm{mem}}_{i,j}))$
                    \EndFor
                    \State $\mathcal{L}_c\gets Q^{-1}\sum_{i=1}^{Q}N_i^{-1}\sum_{j=1}^{N_i}\ell_{i,j}+\lambda\mathcal{R}_1(c)$
                \EndFor
                \State $\widehat{\mathcal{L}}_{\mathrm{comp}}\gets |\mathcal{B}|^{-1}\sum_{c\in\mathcal{B}}
                    \mathcal{L}_c$
                \State $((\phi,\beta),s_{\mathrm{comp}})\gets\operatorname{AdamW}((\phi,\beta),\nabla_{\phi,\beta}\widehat{\mathcal{L}}_{\mathrm{comp}};s_{\mathrm{comp}})$
            \EndFor
        \EndFor
    \end{algorithmic}
\end{algorithm}
\noindent\textbf{Cross-session consolidation.}\quad
Training examples are drawn from the target application's distribution
$\mathcal{D}_{\mathrm{task}}$.
For each example $(\mathcal{C}_{1:T},x,y)$, the frozen session
encoder produces ordered memories $q_1,\ldots,q_T$.  These tensors may be
cached and shared by examples with the same history segments.
We denote the ordered cache for a history by $\mathcal{Q}(\mathcal{C}_{1:T})$,
the configured initial gate by $\psi_0$, and the number of training epochs
by $E_{\mathrm{con}}$.  The operator $\operatorname{Shuffle}$ permutes training
examples using the configured random seed.
\autoref{alg:consolidation} resets the recurrent memory for each example,
initializes it from the first session, and differentiates the task loss
through the subsequent recurrence.  The symbol $\widehat{\ell}$ denotes
the loss of the current example.  Freezing decoder and backbone parameters
preserves differentiation with respect to their memory-conditioned inputs.

\begin{algorithm}[htbp]
    \caption{Application-specific cross-session memory consolidation}
    \label{alg:consolidation}
    \small
    \begin{algorithmic}[1]
        \Require $\mathcal{D}_{\mathrm{task}},E_\xi,G_\phi,D_\beta,f_\theta,\ell_{\mathrm{task}},\psi_0,E_{\mathrm{con}}$
        \Ensure $\psi=\{W_g,b_g\}$
        \State $\psi\gets\psi_0;\ s_{\mathrm{con}}\gets0$
        \For{$(\mathcal{C}_{1:T},x,y)\in\mathcal{D}_{\mathrm{task}}$}
            \State $\mathcal{Q}(\mathcal{C}_{1:T})\gets(G_\phi(E_\xi(c_t)))_{t=1}^{T}$
        \EndFor
        \For{$e=1,\ldots,E_{\mathrm{con}}$}
            \For{$(\mathcal{C}_{1:T},x,y)\in\operatorname{Shuffle}(\mathcal{D}_{\mathrm{task}})$}
                \State $(q_1,\ldots,q_T)\gets\mathcal{Q}(\mathcal{C}_{1:T})$
                \State $h_0\gets0;\ h_1\gets q_1$
                \For{$t=2,\ldots,T$}
                    \State $z_t\gets\sigma([h_{t-1};q_t]W_g+b_g)$
                    \State $h_t\gets z_t\odot h_{t-1}+(1-z_t)\odot q_t$
                \EndFor
                \State $\Lambda_T\gets D_\beta(h_T)$
                \State $\widehat{\ell}\gets\ell_{\mathrm{task}}(p_{\theta\oplus\Lambda_T}(\cdot\mid x),y)$
                \State $(\psi,s_{\mathrm{con}})\gets\operatorname{AdamW}(\psi,\nabla_\psi\widehat{\ell};s_{\mathrm{con}})$
            \EndFor
        \EndFor
    \end{algorithmic}
\end{algorithm}

\noindent\textbf{Online memory and backbone replacement.}\quad
\autoref{alg:deployment} separates arrival of a session from answering a
query.  All module parameters are fixed during deployment.  The retained
memory $h_T$ also serves as the input to an adapted decoder $D_{\beta'}$
for a target backbone $f_{\theta'}$.  Decoder adaptation uses the compilation
objective with target-backbone reference distributions, holding the memory
encoder and target backbone fixed, as described in
\autoref{sec:memory-compiler}.  This adaptation precedes deployment of the
replacement backbone; the accumulated memory, session encoder, and learned
gate are retained across the switch. The target decoder maps the unchanged
memory representation to the new backbone's layer and module dimensions.
The stream $\mathcal{E}$ contains typed events
$\operatorname{Session}(c)$, $\operatorname{Query}(x)$, and
$\operatorname{Replace}(D_{\beta'},f_{\theta'})$.
The operators $\operatorname{Decode}$ and $\operatorname{Emit}$ generate a
response under the chosen decoding policy and return it to the caller,
respectively.  Queries in this algorithm follow the first memory write.

\begin{algorithm}[htbp]
    \caption{Streaming memory updates and query-time decoding}
    \label{alg:deployment}
    \small
    \begin{algorithmic}[1]
        \Require $E_\xi,G_\phi,D_\beta,f_\theta,\psi,\mathcal{E}$
        \State $T\gets0;\ h_0\gets0$
        \For{$e\in\mathcal{E}$}
            \If{$e=\operatorname{Session}(c)$}
                \State $q\gets G_\phi(E_\xi(c))$
                \If{$T=0$}
                    \State $h_{T+1}\gets q$
                \Else
                    \State $z\gets\sigma([h_T;q]W_g+b_g)$
                    \State $h_{T+1}\gets z\odot h_T+(1-z)\odot q$
                \EndIf
                \State $T\gets T+1$
            \ElsIf{$e=\operatorname{Query}(x)$}
                \State $\Lambda_T\gets D_\beta(h_T)$
                \State $y\gets\operatorname{Decode}(p_{\theta\oplus\Lambda_T}(\cdot\mid x))$
                \State $\operatorname{Emit}(y)$
            \ElsIf{$e=\operatorname{Replace}(D_{\beta'},f_{\theta'})$}
                \State $(D_\beta,f_\theta)\gets(D_{\beta'},f_{\theta'})$ \Comment{$h_T,\psi$ unchanged}
            \EndIf
        \EndFor
    \end{algorithmic}
\end{algorithm}
\endgroup

\FloatBarrier
\section{Training Data Construction}
\label{app:training-data}

Compilation trains the memory compiler to encode individual sessions in latent
memory, and consolidation learns a memory-integration policy for a target
application scenario. The two stages use separate data constructions:
query--response pairs for individual sessions, followed by ordered histories
paired with task queries and targets.  Their training procedures are specified
in \appref{app:algorithms}.

\subsection{Single-Session Compilation Data}
\label{app:compilation-data}

\noindent\textbf{Sources and normalization.}\quad
The source mixture covers conversational interactions, task-oriented and
tool-use dialogues, coding-agent trajectories, and synthetic QA.
\autoref{tab:compilation-sources} lists the configured sources and domain-level
sampling targets.  These percentages are token-budget targets used during
construction.  The final corpus is filtered jointly across sources.

\begin{table}[htbp]
    \centering
    \caption{Compilation sources and target shares of context tokens.}
    \label{tab:compilation-sources}
    \small
    \begin{tabular}{lp{0.54\linewidth}c}
        \toprule
        Domain & Sources & Target share \\
        \midrule
        Conversation & UltraChat-200k & 35\% \\
        Task and tool use & Taskmaster-2; Schema-Guided Dialogue; Toucan SFT & 25\% \\
        Coding & SWE-Zero OpenHands trajectories & 20\% \\
        Synthetic QA & D2L synthetic QA, selected level-1 training shards & 20\% \\
        \bottomrule
    \end{tabular}
\end{table}

Source adapters map records to an ordered event stream.  Each event retains
its identifier, speaker role, event type, and content; tool calls additionally
retain available names, arguments, and results.  This representation preserves
the evidence needed to associate each pair with the corresponding interaction.
For Toucan and SWE-Zero, normalization removes system messages, assistant
reasoning fields, and designated reasoning-tool calls.  The synthetic source
is decoded with its source tokenizer before conversion to the same schema.
Long records are segmented at event boundaries under the 4,096-token budgets
of both the Qwen3-8B and ModernBERT tokenizers, with one event of overlap
between adjacent segments.

\noindent\textbf{Filtering.}\quad
Duplicate detection applies Unicode normalization and case folding before
hashing word-token sequences.  Exact duplicates share a SHA-256 hash;
near-duplicate detection uses 64-bit SimHash over three-token shingles,
rejecting signatures within Hamming distance three for texts containing at
least 20 normalized tokens.  Filtering is applied across the accepted corpus
before the training/validation assignment.
The source configuration excludes designated downstream benchmark sources,
and coding trajectories are filtered against an explicit evaluation-repository
exclusion list.  Source identifiers, revisions, and record provenance are
retained in the data manifests.

\autoref{tab:source-revisions} lists the repositories and revision prefixes
specified in the source configuration.  The coding exclusion list contains
the 100 repositories represented by the 500 tasks in the verified split of
\nolinkurl{SWE-bench-Live/SWE-bench-Live}, revision
\texttt{a637bd46829f}.  This repository-level filter is applied during source
normalization.  The synthetic subset selects level-1 training shards and
decodes their token sequences with the Mistral-7B-Instruct-v0.2 tokenizer.

\begin{table}[htbp]
    \centering
    \caption{Configured compilation data sources. Revision identifiers are abbreviated to 12 characters.}
    \label{tab:source-revisions}
    \small
    \begin{tabular}{p{0.69\linewidth}l}
        \toprule
        Repository and subset & Revision \\
        \midrule
        \nolinkurl{HuggingFaceH4/ultrachat_200k} & \texttt{8049631c405a} \\
        \nolinkurl{DeepPavlov/TaskMaster2} & \texttt{333d425dd1f5} \\
        \nolinkurl{google-research-datasets/}\newline
        \nolinkurl{dstc8-schema-guided-dialogue}, training dialogues & \texttt{e852981ae349} \\
        \nolinkurl{Agent-Ark/Toucan-1.5M}, SFT subset & \texttt{0df3cf37f2ab} \\
        \nolinkurl{nvidia/SWE-Zero-openhands-trajectories} & \texttt{7b3cd106d00f} \\
        \nolinkurl{SakanaAI/self_gen_qa_d2l}, level-1 training shards & Selected shards \\
        \bottomrule
    \end{tabular}
\end{table}

\noindent\textbf{Query--response pairs.}\quad
Every accepted session has exactly ten query--response pairs.
A pair contains a query and a fixed reference response; answerable pairs
also cite supporting session events.
Existing source-grounded pairs are preserved, and missing pairs are
generated and validated against the same evidence-linking contract.
For answerable questions, the generation prompt requests concise responses
grounded in explicit event content, with evidence identifiers copied from the session.
The preparation pipeline also supports source-derived response pairs and
deterministic event-recall fallbacks to complete pair coverage.
Validation checks the pair schema, nonempty reference responses, answerability
flags, evidence identifiers, exact pair counts, and preservation of the
underlying session content.  These checks establish structural and evidence-link
consistency; reference responses remain the fixed trajectories used in
distribution matching.

\autoref{tab:probe-generation} specifies the documented reproduction
configuration for query--response generation. The generator returns a compact JSON
record containing the query, question type, answerability flag, reference
response, and evidence-event identifiers.  Answerable pairs must cite valid
events; the protocol also permits unanswerable pairs with an empty evidence
list. All retained pairs receive the same training treatment, without
answerability-based filtering or weighting at training time. The parser removes
duplicate query--response pairs and validates each accepted pair against its
session.  Generation continues for sessions with incomplete coverage, up to
the configured round limit.  Event-grounded fallbacks then complete the ten
pairs, and a final per-source validation checks coverage and preservation
of session content.  Source-derived and fallback pairs follow the same
schema as generated pairs.
Prompt examples are provided in
\appref{app:prompt-examples}.

\begin{table}[htbp]
    \centering
    \caption{Documented query--response generation configuration.}
    \label{tab:probe-generation}
    \small
    \begin{tabular}{ll}
        \toprule
        Setting & Value \\
        \midrule
        Generator & Locally hosted Qwen3-8B \\
        Generation policy & \texttt{grounded\_compact\_v3} \\
        Pairs per session & 10 \\
        Temperature & 0 \\
        Maximum generated tokens per request & 768 \\
        Generation batch size & 64 \\
        Maximum generation rounds & 3 \\
        Additional source-derived pairs per session & 0 by default \\
        Event-grounded fallback & Enabled \\
        \bottomrule
    \end{tabular}
\end{table}

\noindent\textbf{Corpus freeze and splits.}\quad
The resulting frozen corpus contains 671,030 sessions, 6,710,300 pairs, and
525,898,190 backbone-tokenized context tokens.
Its deterministic split assigns eight pairs from each of 664,128 sessions to
training, reserves two disjoint pairs from those sessions for query-level
validation, and holds out another 6,902 sessions for session-level validation.
The session-level assignment uses a seeded hash of the session identifier,
with a 1\% holdout probability.  Within training sessions, a second seeded
hash orders pair identifiers to select the two query-held-out pairs.
Both assignments use seed 42.  Session-level validation evaluates unseen
session content, whereas query-level validation evaluates disjoint pairs
over training-session content.

\autoref{tab:compilation-splits} gives the pair counts for each partition.
Training and query-level validation share session content and use disjoint
pair sets; session-level validation is disjoint in session content.
\begin{table}[htbp]
    \centering
    \caption{Compilation data partitions.}
    \label{tab:compilation-splits}
    \small
    \begin{tabular}{lrrr}
        \toprule
        Partition & Sessions & Pairs per session & Total pairs \\
        \midrule
        Training & 664,128 & 8 & 5,313,024 \\
        Query-level validation & 664,128 & 2 & 1,328,256 \\
        Session-level validation & 6,902 & 10 & 69,020 \\
        \bottomrule
    \end{tabular}
\end{table}

\noindent\textbf{Reference probability targets.}\quad
For every fixed reference response, the frozen base language model is evaluated
with the corresponding session present in its textual context.
At each response position, the data pipeline stores the model's top-$K$ token
log-probabilities; the remaining mass is reconstructed as the tail category in
\autoref{eq:topk-tail-projection}.
The compiler is trained from these offline probability targets using
\autoref{eq:compilation-fkl}.
The adapter-conditioned model receives the same query and response prefix,
with historical information supplied through the generated adapter.
Only response positions contribute to distribution matching.  Averaging first
within each pair and then within each session gives equal session weight
despite variable response lengths.  Reference support size, sequence limits,
and optimizer settings are specified in \appref{app:compiler-configuration}.

\subsection{Cross-Session Consolidation Data}
\label{app:consolidation-data}

\noindent\textbf{Training-example construction.}\quad
Each downstream example comprises the chronological history available at its
query boundary, the query, candidate answers, and the correct answer label.
The history is converted to bounded inputs using the compilation-stage event
representation.  Native session boundaries are preserved where available;
long sessions are divided into chronological segments with a 4,096-token
budget and one-event overlap.  Oversized events are split into token chunks
to retain their content.  The session encoder consumes the history segments,
while the query and candidate answers are supplied to the answering backbone.
The correct label supplies task supervision.  Frozen session encodings are
cached with their order and segmentation metadata so that multiple questions
can reuse the same encoded history.

\noindent\textbf{\permabench.}\quad
For every evaluated history variant, each question is paired with its
query-specific visible session prefix and official A--H candidate set.
Ten leave-one-user-out folds assign nine users to gate training and the
remaining user to evaluation.  All three task types of the training users
contribute supervision.  Each example starts a fresh recurrence over its
visible history, with the first segment initializing memory and subsequent
segments processed in chronological order.  The held-out user's history is
encoded at evaluation using the fixed compiler and trained gate.

\noindent\textbf{\personamembench.}\quad
Construction starts from the pinned 32K-Text release and its training,
validation, and benchmark partitions.  Training and validation rows belonging
to benchmark personas are excluded.  Two known invalid training rows are
removed: one has no distractor answers and one includes the correct answer
among the distractors.  Null distractor entries are discarded with the repair
recorded.  The resulting partitions contain 18,527 training examples, 2,059
validation examples, and all 5,000 benchmark questions.
Each example resolves its linked chat history into a chronological message
stream, which is segmented under the shared token budget.  Candidate answers
are deterministically shuffled using the instance identifier, and the same
ordering is used by every method.  Training uses the training partition;
the final-epoch gate is evaluated on the benchmark partition.

\noindent\textbf{\prefevalbench.}\quad
Each base example is represented in three evidence forms: an explicit
statement, a choice-based interaction, and a persona-based interaction.
The corresponding evidence sessions are followed by the official intervening
dialogue prefix for each of the 10-, 70-, and 300-turn settings.
The frozen preparation contains 1,000 base examples and 3,000 form-specific
examples before expansion over these three history lengths.
The topic split follows the official training implementation with seed 42:
16 topics are used for training, while transportation, technology shopping,
education resources, and motor shopping are held out.  One gate is trained
jointly across training topics, evidence forms, and history lengths.
The split contains 820 training and 180 held-out base examples, yielding
2,460 and 540 form-specific examples, respectively.  Expansion over the
three history lengths produces 7,380 training instances and 1,620 evaluation
instances; each history-length setting evaluates the same 540 held-out
form-specific examples.  Base examples and topics are disjoint across the
training and evaluation partitions.
Shared intervening sessions reuse their cached encodings.  Answer options
are deterministically permuted with seed 42 and the instance identifier,
consistently across methods.

\noindent\textbf{Supervision and optimization.}\quad
For these multiple-choice experiments, the task loss in
\autoref{eq:phase2-objective} supplies cross-entropy supervision on the
correct answer label after option permutation. On \permabench{}, training
uses the full-vocabulary next-token logits with the correct answer-label
token as the target; evaluation selects the highest-logit legal option.
Gradients pass through the frozen answering backbone and decoder to the
recurrent gate. Across training examples, this supervision learns a shared
integration policy for the application's memory requirements. The policy is
reused across subsequent interactions with its parameters fixed after deployment.
Training parameters and checkpoint
selection are specified in \appref{app:compiler-configuration}; evaluation
metrics and aggregation are specified in \appref{app:benchmark-protocols}.

\FloatBarrier
\Needspace{12\baselineskip}
\subsection{Query--Response Generation Prompts}
\label{app:prompt-examples}

The following prompt examples illustrate grounded query--response generation
from session histories. Wording and output-format notation are simplified
for readability while preserving the generation requirements.

\newtcolorbox{paperprompt}[1]{
    enhanced,
    breakable,
    colback=gray!10,
    colframe=gray!50!black,
    colbacktitle=gray!50!black,
    coltitle=white,
    fonttitle=\bfseries,
    fontupper=\small,
    title={#1},
    boxrule=0.7pt,
    arc=2mm,
    left=9pt,right=9pt,top=10pt,bottom=8pt,
    before skip=14pt,after skip=12pt,
    attach boxed title to top center={yshift=-2mm},
    boxed title style={sharp corners=north,colframe=gray!50!black},
    before upper={\setlength{\parindent}{0pt}\setlength{\parskip}{5pt}}
}

\begin{paperprompt}{Query--Response Generation: System Prompt}
Create concise question--answer pairs grounded in the provided agent session.

\textbf{Requirements}
\begin{enumerate}[leftmargin=1.5em,itemsep=3pt,topsep=3pt]
    \item Use only facts explicitly present in the session history.
    \item For each answerable question, provide a reference answer and cite
    one or more supporting events. Copy event identifiers exactly from the
    history; do not invent identifiers or cite the session as a whole.
    \item Mark unanswerable questions accordingly and leave their supporting
    event list empty.
    \item Keep questions and reference answers concise.
\end{enumerate}

\textbf{Output}

Return one compact JSON object containing each question, its type,
answerability, reference answer, and supporting event identifiers.
Do not include additional prose, Markdown, memory summaries, or reasoning.
\end{paperprompt}

\begin{paperprompt}{Query--Response Generation: User Prompt}
Generate the remaining grounded question--answer pairs so that the session
has exactly ten pairs in total. Do not repeat any existing questions.

\textbf{Existing questions, if any:}\quad [Existing questions]

\textbf{Session history:}\quad [Session events with their identifiers]
\end{paperprompt}

\FloatBarrier
\section{Ablation Details}
\label{app:ablation-details}

\subsection{Compilation Objective Variants}
\label{app:compilation-objectives}

The objective ablation in \autoref{tab:objective-ablation} varies two
properties of compiler training: the response trajectory used to evaluate the
token distributions and the divergence defined on those distributions.
We use the notation introduced in \autoref{sec:memory-compiler}.
For pair $i$ and response position $j$, $p^{\mathrm{ctx}}_{i,j}$ is the
next-token distribution produced when the frozen language model receives the
session as textual context, and $p^{\mathrm{mem}}_{i,j}$ is the distribution
produced when the same model receives the compiler-generated adapter.
The fixed response token at this position is $y_{i,j}$, and $S_{i,j}$ contains
the $K$ highest-probability tokens under $p^{\mathrm{ctx}}_{i,j}$.

\paragraph{Fixed-trajectory objectives.}
\compilersft{} treats each fixed reference response as a sequence of hard token
targets.  Its token-level loss is
\begin{equation}
    \ell^{\mathrm{SFT}}_{i,j}
    =-\log p^{\mathrm{mem}}_{i,j}(y_{i,j}).
    \label{eq:compiler-sft-variant}
\end{equation}
Here, $\ell^{\mathrm{SFT}}_{i,j}$ is the negative log-likelihood of reference
token $y_{i,j}$ under the adapter-conditioned distribution.
Top-$K$ CE replaces the hard target with the probabilities of the selected
context-conditioned tokens:
\begin{equation}
    \ell^{\mathrm{TopK}}_{i,j}
    =-\sum_{v\in S_{i,j}}p^{\mathrm{ctx}}_{i,j}(v)
      \log p^{\mathrm{mem}}_{i,j}(v).
    \label{eq:compiler-topk-ce-variant}
\end{equation}
The symbol $v$ indexes a token in the shared vocabulary, and
$\ell^{\mathrm{TopK}}_{i,j}$ denotes the selected-support cross-entropy.
\fixedfkl{} uses the same fixed response trajectory and selected support, then
represents all tokens outside $S_{i,j}$ by their aggregate probability mass.
Its token-level loss is defined in \autoref{eq:topk-tail-projection}, and the
session-balanced aggregation and adapter regularization follow
\autoref{eq:session-fkl} and \autoref{eq:compilation-fkl}.

\paragraph{Sampled-trajectory objectives.}
The sampled-trajectory variants sample a response
$\widetilde{y}_i=(\widetilde{y}_{i,1},\ldots,
\widetilde{y}_{i,\widetilde{N}_i})$ from the adapter-conditioned model, where
$\widetilde{N}_i$ is its generated length and $\widetilde{y}_{i,<j}$ is the
prefix preceding position $j$.
Both models are then evaluated along this shared sampled prefix:
\begin{equation}
    \widetilde{p}^{\mathrm{ctx}}_{i,j}(v)
    =p_\theta(v\mid c,x_i,\widetilde{y}_{i,<j}).
    \label{eq:online-context-distribution}
\end{equation}
\begin{equation}
    \widetilde{p}^{\mathrm{mem}}_{i,j}(v)
    =p_{\theta\oplus\Lambda(c)}(v\mid x_i,\widetilde{y}_{i,<j}).
    \label{eq:online-memory-distribution}
\end{equation}
The tilde distinguishes distributions evaluated on the sampled trajectory from
the fixed-trajectory distributions in \autoref{eq:context-distribution} and
\autoref{eq:compilation-distributions}.
\sampledfkl{} applies the forward-KL construction in
\autoref{eq:topk-tail-projection} to
$\widetilde{p}^{\mathrm{ctx}}_{i,j}$ and
$\widetilde{p}^{\mathrm{mem}}_{i,j}$.

\sampledrkl{} selects the top-$K$ support
$\widetilde{S}_{i,j}$ under $\widetilde{p}^{\mathrm{mem}}_{i,j}$ and reverses
the order of the two distributions.
For $a\in\{\mathrm{ctx},\mathrm{mem}\}$, let
$\widetilde{\tau}^{a}_{i,j}$ denote the probability mass that
$\widetilde{p}^{a}_{i,j}$ assigns outside $\widetilde{S}_{i,j}$.
Its token-level objective is
\begin{equation}
    \ell^{\mathrm{RKL}}_{i,j}
    =\sum_{v\in\widetilde{S}_{i,j}}
      \widetilde{p}^{\mathrm{mem}}_{i,j}(v)
      \log\frac{\widetilde{p}^{\mathrm{mem}}_{i,j}(v)}
                    {\widetilde{p}^{\mathrm{ctx}}_{i,j}(v)}
      +\widetilde{\tau}^{\mathrm{mem}}_{i,j}
      \log\frac{\widetilde{\tau}^{\mathrm{mem}}_{i,j}}
                    {\widetilde{\tau}^{\mathrm{ctx}}_{i,j}}.
    \label{eq:online-rkl-variant}
\end{equation}
Here, $\ell^{\mathrm{RKL}}_{i,j}$ is the top-$K$-plus-tail approximation to
the reverse KL at the sampled response position.

\sampledrkl{} (K3) estimates the same divergence from the sampled response token.
We define the sampled-token log-ratio as
\begin{equation}
    r_{i,j}
    =\log\widetilde{p}^{\mathrm{ctx}}_{i,j}(\widetilde{y}_{i,j})
     -\log\widetilde{p}^{\mathrm{mem}}_{i,j}(\widetilde{y}_{i,j}).
    \label{eq:k3-log-ratio}
\end{equation}
The corresponding K3 loss is
\begin{equation}
    \ell^{\mathrm{K3}}_{i,j}=\exp(r_{i,j})-r_{i,j}-1,
    \label{eq:online-k3-variant}
\end{equation}
where $r_{i,j}$ compares the two models' log-probabilities for sampled token
$\widetilde{y}_{i,j}$ and $\ell^{\mathrm{K3}}_{i,j}$ is its nonnegative
single-sample reverse-KL estimator.
All variants average token losses within pairs and then across pairs, and all
use the same adapter regularization as \autoref{eq:compilation-fkl}.
Each sampled-trajectory run begins with 10,377 SFT warm-start updates,
then switches to its respective objective for the remaining 41,508 updates
within the same run, reaching the common 51,885-update budget.
Every trained compiler is frozen before fitting an independent
\consolidationgate{} with the same downstream architecture and optimization
protocol.

\begin{table}[htbp]
    \centering
    \caption{Effect of compiler training objectives on \permabench{} accuracy
    (\%).}
    \label{tab:objective-ablation}
    \small
    \setlength{\tabcolsep}{4.2pt}
    \begin{tabular}{lccccc}
        \toprule
        Objective & Clean SD & Noisy SD & Clean MD & Noisy MD & Avg. \\
        \midrule
        \compilersft{} & 72.43 & 71.97 & 50.47 & 49.98 & 61.21 \\
        \sampledrkl{} (top-$K$) & 73.53 & 74.63 & 60.49 & 58.97 & 66.91 \\
        \sampledrkl{} (K3) & 71.03 & 70.60 & 63.42 & 63.02 & 67.02 \\
        Top-$K$ CE & 82.10 & 83.87 & 73.30 & 71.37 & 77.66 \\
        \sampledfkl{} & 88.90 & 87.83 & 79.23 & 79.04 & 83.75 \\
        \textbf{\fixedfkl{}} & \textbf{90.57} & \textbf{91.20} &
        \textbf{79.67} & \textbf{80.62} & \textbf{85.52} \\
        \bottomrule
    \end{tabular}
\end{table}

\subsection{Compilation Training Dynamics}
\label{app:compiler-dynamics}

\begin{figure}[htbp]
    \centering
    \includegraphics[width=\linewidth]{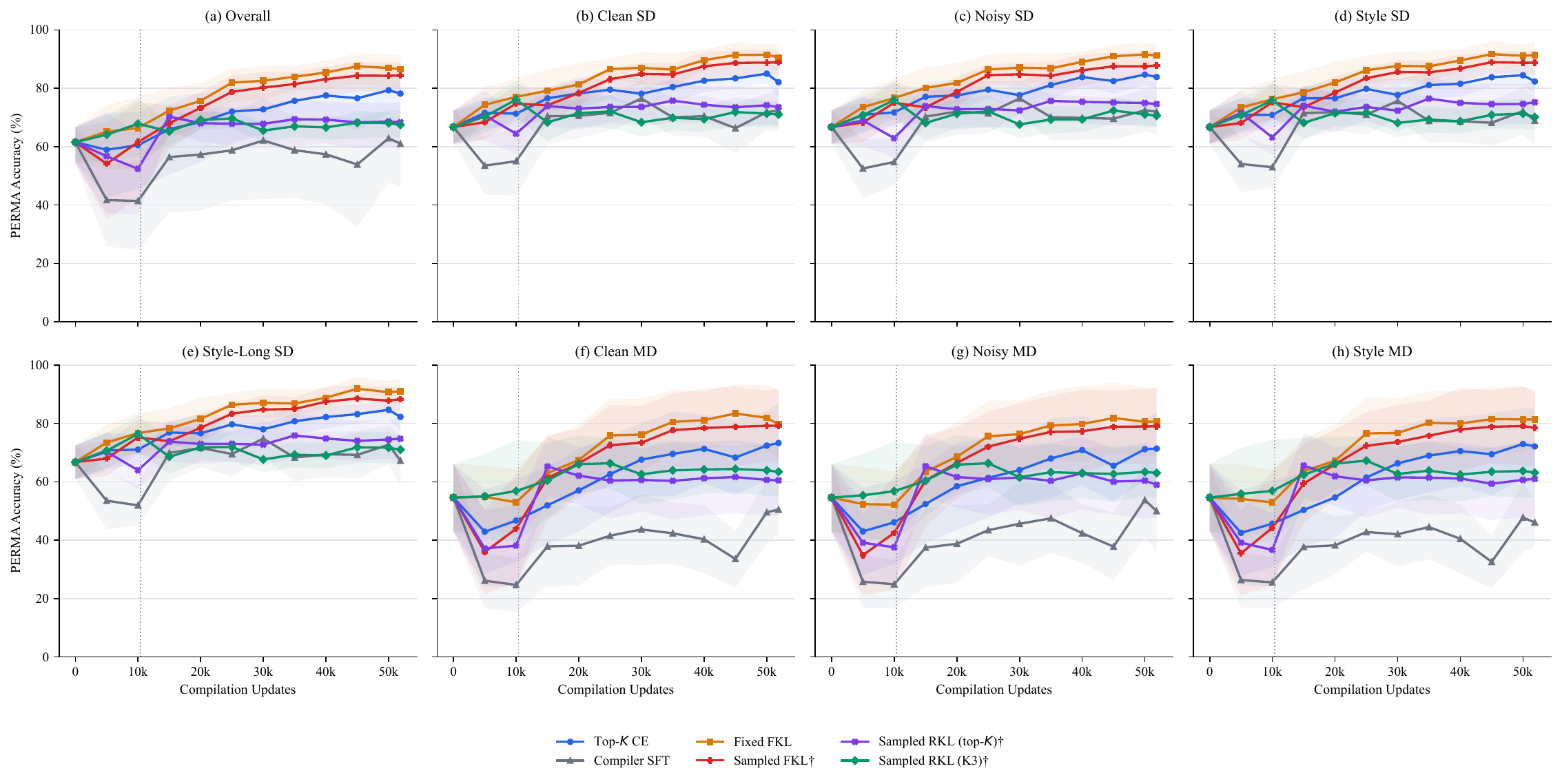}
    \caption{Compilation training dynamics on \permabench{}.
    Panel (a) reports the unweighted average over all seven settings; its
    shaded regions span the minimum and maximum setting-level accuracy.
    Panels (b)--(h) report each setting separately; their shaded regions show
    the standard deviation across the ten held-out-user folds.
    Every checkpoint is evaluated after independently fitting the same
    \consolidationgate{} architecture. Daggered variants switch to their
    respective objectives after 10,377 SFT warm-start updates within each run.}
    \label{fig:compiler-dynamics}
\end{figure}

We evaluate intermediate compilation checkpoints by freezing each compiler and
fitting an independent \consolidationgate{} with the same downstream protocol.
\autoref{fig:compiler-dynamics} reports both the average trajectory over all
seven \permabench{} settings and the corresponding trajectory within each
setting.
\fixedfkl{} establishes the highest average downstream accuracy by
approximately 15,000 updates and preserves this lead through the
51,885-update budget.
The same ordering appears throughout the seven individual settings, while
\compilersft{} exhibits an early decrease followed by a gradual recovery.
These trajectories show that the endpoint comparison in
\autoref{tab:objective-ablation} reflects a stable difference in compilation
quality over training.

\subsection{Cross-Session Consolidation Variants}
\label{app:consolidation-variants}

The consolidation ablation in \autoref{tab:fusion-ablation} fixes the
\fixedfkl{} compiler and changes the operation applied to the chronologically
ordered sequence of memory representations.  All variants share the compiler
checkpoint, input segmentation, visible-history boundary, held-out-user folds,
and evaluation protocol.  The compiler and backbone remain frozen; the learned
rule fits the consolidation gate on the training users of each fold.

\noindent\textbf{Factor convention.}\quad
Let $S$ denote the number of compiler segments in the visible history and
$s\in\{1,\ldots,S\}$ index their chronological order.  For one adapted layer
and module, suppressing the indices $\ell,m$, let
$A_s\in\mathbb{R}^{r\times d^{\mathrm{in}}}$ and
$B_s\in\mathbb{R}^{d^{\mathrm{out}}\times r}$ be the decoded factors of
segment $s$.  Here $r$ is the generated factor width, and
$d^{\mathrm{in}},d^{\mathrm{out}}$ are the target module's input and output
dimensions.  The decoder also supplies a shared factor pair $A_0,B_0$, called
the head bias in the implementation, which contributes the history-independent
update $\Delta W_0=\gamma B_0A_0$.  The coefficient $\gamma$ is the checkpoint's
LoRA scaling from \autoref{eq:generated-lora}.  With head bias enabled, this
shared pair occupies one additional rank-$r$ block.

\noindent\textbf{Selection and latent averaging.}\quad
Latest Session retains all compiler segments belonging to the final
observed semantic session and combines their decoded factors by rank
concatenation.  Latent averaging takes the element-wise mean of all $S$
segment representations and decodes that single mean.  Learned consolidation
instead decodes the accumulated memory produced by
\mbox{\autoref{sec:memory-consolidator}}.  Both latent averaging and learned
consolidation produce one rank-$r$ generated block plus the shared bias block.

\noindent\textbf{Rank concatenation.}\quad
Concatenation stacks the down-projection factors vertically and the
up-projection factors horizontally, appending the shared bias pair once:
\begin{align}
    A_{\mathrm{cat}}&=[A_1^\top,\ldots,A_S^\top,A_0^\top]^\top,
    \label{eq:concat-a}\\
    B_{\mathrm{cat}}&=[B_1,\ldots,B_S,B_0],
    \label{eq:concat-b}\\
    \Delta W_{\mathrm{cat}}
    &=\gamma B_{\mathrm{cat}}A_{\mathrm{cat}}
      =\gamma\sum_{s=1}^{S}B_sA_s+\Delta W_0.
    \label{eq:concat-update}
\end{align}
The resulting factor width is $(S+1)r$ with head bias enabled; Latest Session
uses the same construction over the selected session's segments.
The update is an unnormalized sum under the unchanged coefficient $\gamma$.

\noindent\textbf{LoRA factor averaging.}\quad
This rule computes the arithmetic mean of each generated factor separately:
\begin{align}
    \overline A&=\frac{1}{S}\sum_{s=1}^{S}A_s,
    \label{eq:factor-mean-a}\\
    \overline B&=\frac{1}{S}\sum_{s=1}^{S}B_s,
    \label{eq:factor-mean-b}\\
    \Delta W_{\mathrm{fac}}
    &=\gamma\overline B\,\overline A+\Delta W_0
      =\frac{\gamma}{S^2}\sum_{s=1}^{S}\sum_{u=1}^{S}B_sA_u+\Delta W_0.
    \label{eq:factor-mean-update}
\end{align}
Here $u$ independently indexes the down-projection factor in the expanded
product.  The shared bias factors are identical across segments and remain
unchanged by averaging, giving a total factor width of $2r$.  Thus this
operator includes cross-segment products $B_sA_u$.  Averaging the induced
updates is a distinct operator, $\gamma S^{-1}\sum_s B_sA_s+\Delta W_0$;
the reported row evaluates factor averaging.

The variants in \autoref{tab:fusion-ablation} retain the same checkpoint scaling $\gamma$, applied directly to
the factor product without division by the assembled factor width.  Update
norms follow from the specified operators, with no additional post-aggregation
norm calibration.  Concatenation has history-dependent factor width, whereas
latent averaging, factor averaging, and learned consolidation have fixed factor width.

\begin{table}[htbp]
    \centering
    \caption{Cross-session consolidation ablation with the \fixedfkl{}
    compiler on \permabench{} (accuracy, \%).}
    \label{tab:fusion-ablation}
    \small
    \setlength{\tabcolsep}{4.2pt}
    \begin{tabular}{lccccc}
        \toprule
        Consolidation & Clean SD & Noisy SD & Clean MD & Noisy MD & Avg. \\
        \midrule
        Rank concatenation & 14.83 & 15.10 & 12.70 & 12.71 & 13.84 \\
        Latent averaging & 63.13 & 61.67 & 28.67 & 28.61 & 45.52 \\
        Latest Session & 68.07 & 67.37 & 38.96 & 39.83 & 53.56 \\
        Factor averaging & 69.50 & 69.37 & 38.99 & 38.99 & 54.21 \\
        \textbf{Learned gate (ours)} & \textbf{90.57} &
        \textbf{91.20} & \textbf{79.67} & \textbf{80.62} & \textbf{85.52} \\
        \bottomrule
    \end{tabular}
\end{table}

\FloatBarrier
\phantomsection
\noindent\textbf{Additional consolidation experiments.}
\label{app:additional-consolidation}
\par\nopagebreak[4]

\paragraph{Experimental setup.}
We evaluate alternative consolidation rules on PERMA Clean SD and Clean MD
using Qwen3-8B and the same frozen five-pass memory compiler. Each setting
uses ten held-out-user folds, with nine users available for training or
coefficient selection and the remaining user reserved for evaluation.
All methods share the compiler, decoder, backbone, history segmentation,
and evaluation questions. Histories are segmented at 4,096 tokens with
one-message overlap; updates operate on these compiler segments.
Latent averaging and RPMem are rerun alongside the additional controls.
Accuracy is macro-averaged over the ten held-out users, and the two-setting
mean gives equal weight to Clean SD and Clean MD.

\paragraph{Aggregation rules.}
Latent averaging takes the arithmetic mean of the segment representations
before decoding. Tuned exponential moving averaging (EMA) initializes
$h_1=q_1$ and applies $h_t=\alpha h_{t-1}+(1-\alpha)q_t$ thereafter.
For each fold and setting, we select $\alpha$ from
$\{0,0.25,0.5,0.75,0.9,0.95,0.99\}$ using macro-average accuracy on the
nine training-side users, with ties resolved toward the smaller coefficient.
The selected coefficient is fixed for the held-out user.
Update averaging decodes each segment separately and averages its induced
weight update:
\[
  \Delta W=\frac{\gamma}{S}\sum_{s=1}^{S}B_sA_s+\Delta W_0,
\]
where $S$ is the number of segments, $\gamma$ is the compiler's LoRA scale,
and $\Delta W_0$ is the shared decoder bias update, included once.
This rule averages complete updates rather than the two factors separately.
Its factor-concatenation implementation retains a rank that grows with $S$.

\paragraph{Learned consolidation.}
The scalar gate mean-pools the previous memory and incoming representation
over their layer, module, and rank axes, concatenates the resulting two
512-dimensional vectors, and applies a learned linear map and sigmoid.
The resulting retention coefficient is shared across all memory coordinates;
the gate has 1,025 parameters. RPMem instead uses coordinate-dependent
retention. Both learned methods initialize $h_1=q_1$ and apply consolidation
from the second segment onward. Both are trained on the other nine users
with the same answer-token cross-entropy objective for five epochs, using
AdamW with learning rate $10^{-3}$, weight decay $0.01$, gradient clipping
at $1.0$, and seed 42. The compiler and backbone remain frozen, and neither
gate receives the evaluation question as an input.

\begin{table}[H]
\centering
\caption{Additional consolidation experiments on PERMA (accuracy, \%).
Each setting averages ten held-out-user folds. Mean averages the two displayed
settings; it is distinct from the four-setting Core Avg. in the main results.
All values are from the matched evaluation runs described here.}
\label{tab:additional-consolidation}
\small
\setlength{\tabcolsep}{4pt}
\begin{tabular*}{\linewidth}{@{\extracolsep{\fill}}lcccc@{}}
\toprule
Method & Task supervision & Clean SD & Clean MD & Mean \\
\midrule
Latent averaging & None & 63.13 & 28.67 & 45.90 \\
Update averaging & None & 68.57 & 38.12 & 53.34 \\
Scalar gate & Answer-token training & 67.57 & 39.21 & 53.39 \\
Tuned EMA & Coefficient selection & 68.07 & 38.96 & 53.51 \\
RPMem & Answer-token training & \textbf{90.80} & \textbf{79.13} & \textbf{84.97} \\
\bottomrule
\end{tabular*}
\end{table}

\paragraph{Results.}
Update averaging and tuned EMA achieve two-setting means of 53.34\% and
53.51\%, respectively, compared with 45.90\% for latent averaging.
EMA selects $\alpha=0$ in all twenty setting--fold combinations, favoring
the most recent segment within the evaluated coefficient family.
The learned scalar gate reaches 53.39\%, whereas RPMem reaches 84.97\%.
Under the same task supervision and training budget, coordinate-dependent
retention therefore substantially outperforms a single shared retention
coefficient on both settings. These comparisons support the value of
fine-grained recurrent consolidation beyond uniform averaging and
globally weighted updates.

\subsection{Amount of Consolidation Supervision}
\label{app:consolidation-supervision}

\begin{figure}[H]
    \centering
    \includegraphics[width=0.76\linewidth]{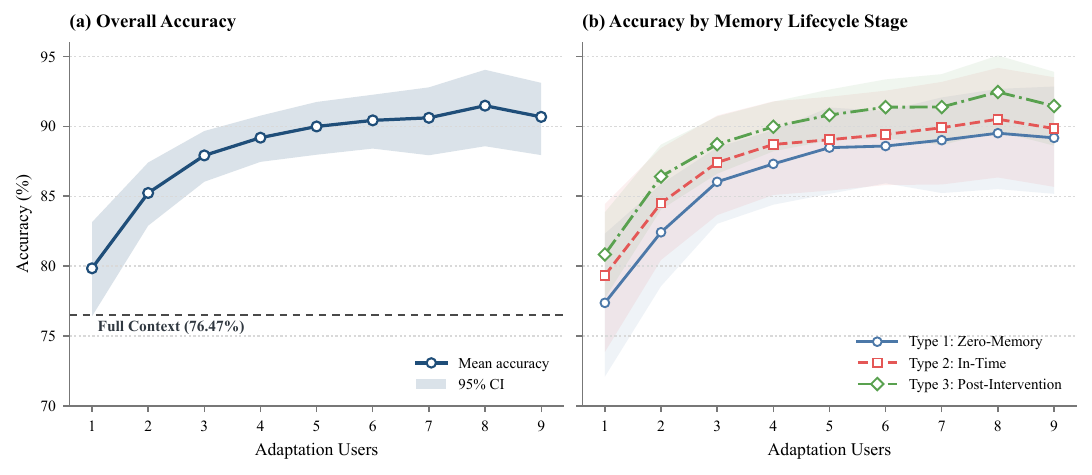}
    \caption{Data efficiency of consolidation adaptation on \permabench{}
    Clean SD. Curves report held-out users, and shaded regions show user-level
    bootstrap 95\% confidence intervals. The horizontal reference marks
    Full Context at 76.47\%.}
    \label{fig:gate-data-efficiency}
\end{figure}

We vary the number of training users from one to nine and evaluate each gate
on users excluded from its training set.
For every training-set size, cyclic sampling produces ten balanced
train-evaluation partitions, and results are first aggregated within users.
As shown in \autoref{fig:gate-data-efficiency}, one-user adaptation reaches
79.84\% on Clean SD, and five-user adaptation reaches 89.98\%.
Accuracy remains between 89.98\% and 91.47\% from five to nine users.
The learning curve demonstrates that the consolidation policy transfers across
users with limited downstream supervision.

\FloatBarrier
\begingroup
\subsection{Comparison with SHINE and Recurrent Consolidation}
\label{sec:additional-shine}

We evaluate an existing context-to-LoRA compiler with and without learned recurrent consolidation on PERMA. All methods use Qwen3-8B and the same ten-fold leave-one-user-out evaluation across seven PERMA variants. SHINE uses the publicly released \texttt{SHINE-ift\_mqa} checkpoint. SHINE w/o Gate applies fixed latent averaging to the compiled session representations before decoding. SHINE w/ Gate freezes the SHINE compiler and decoder and trains the same recurrent consolidation gate architecture used by RPMem for five epochs using the nine training users in each fold. RPMem w/ Gate uses the fixed-reference forward-KL compiler and the matched downstream gate-training protocol.

\begin{table}[htbp]
\centering
\caption{Comparison with SHINE on PERMA (accuracy, \%). Core Avg. averages the four clean/noisy SD/MD settings; 7-Variant Avg. averages all seven variants.}
\label{tab:additional-shine}
\small
\begin{tabular}{lrr}
\hline
Method & Core Avg. & 7-Variant Avg. \\
\hline
SHINE w/o Gate & 28.64 & 29.26 \\
SHINE w/ Gate & 84.25 & 85.01 \\
RPMem w/ Gate & \textbf{85.52} & \textbf{86.53} \\
\hline
\end{tabular}
\end{table}

\begin{table}[htbp]
\centering
\caption{Setting-level SHINE comparison on PERMA (accuracy, \%).}
\label{tab:additional-shine-full}
\small
\begin{tabular}{lrrr}
\hline
Setting & SHINE w/o Gate & SHINE w/ Gate & RPMem w/ Gate \\
\hline
Clean SD & 34.00 & \textbf{92.13} & 90.57 \\
Noisy SD & 33.47 & 91.10 & \textbf{91.20} \\
Style SD & 33.30 & \textbf{91.77} & 91.37 \\
Style-Long SD & 33.40 & \textbf{91.23} & 90.97 \\
Clean MD & 23.13 & 76.81 & \textbf{79.67} \\
Noisy MD & 23.97 & 76.97 & \textbf{80.62} \\
Style MD & 23.54 & 75.04 & \textbf{81.33} \\
\hline
SD Avg. & 33.54 & \textbf{91.56} & 91.03 \\
MD Avg. & 23.55 & 76.27 & \textbf{80.54} \\
7-Variant Avg. & 29.26 & 85.01 & \textbf{86.53} \\
Core Avg. & 28.64 & 84.25 & \textbf{85.52} \\
\hline
\end{tabular}
\end{table}

Adding the learned recurrent gate raises SHINE's seven-variant accuracy from 29.26\% to 85.01\% (+55.75 percentage points). With the same gate architecture and downstream supervision, RPMem reaches 86.53\%, a further gain of 1.52 points; the core-setting advantage is 1.27 points. These results distinguish the contribution of learned consolidation from the additional gains of RPMem's session compilation and latent representation in this combined setting.

\subsection{Zero-Shot Transfer of Recurrent Consolidation}
\label{sec:additional-gate-transfer}

We evaluate whether a consolidation gate trained on PERMA can be reused on PersonaMem-v2 without target-benchmark gate training. RPMem uses Qwen3.5-9B. We transfer 70 PERMA-trained gates obtained from seven PERMA variants and ten leave-one-user-out folds per variant. The session compiler, model-specific decoder, and gates are frozen. Every gate is evaluated on all 5,000 PersonaMem-v2 questions. Accuracies are averaged over the ten source folds within each PERMA variant and then equally across the seven variants.

\begin{table}[htbp]
\centering
\caption{Cross-benchmark transfer to PersonaMem-v2 (accuracy, \%).}
\label{tab:additional-gate-transfer}
\small
\begin{tabular}{llr}
\hline
Method & Gate training source & Accuracy \\
\hline
RPMem, latent averaging & None & 35.25 \\
Metis-9B, released & Released checkpoint & 29.64 \\
RPMem, recurrent gate & PERMA & 38.14 \\
RPMem, recurrent gate & PersonaMem-v2 & \textbf{45.73} \\
\hline
\end{tabular}
\end{table}

The PERMA-trained gates achieve 38.14\% without PersonaMem-v2 gate adaptation, exceeding latent averaging by 2.89 percentage points and the released Metis-9B checkpoint by 8.50 points. Training the gate directly on PersonaMem-v2 raises accuracy to 45.73\%, illustrating the additional benefit of target-task supervision.

\begingroup
\noindent\textbf{Backbone-matched comparison with textual memory.}\quad
To compare zero-shot consolidation transfer with textual memory methods under the same backbone, we additionally evaluated RPMem on PersonaMem-v2 using Qwen3-8B and PERMA-trained consolidation gates. The memory compiler, decoder, and gates were frozen, without any PersonaMem-v2-specific supervised training. The textual baselines likewise require no target-benchmark supervised training. All methods use the same 5,000-question PersonaMem-v2 test set and evaluation protocol.

\begin{table}[htbp]
\centering
\caption{Backbone-matched zero-shot comparison on PersonaMem-v2 (Overall accuracy, \%).}
\label{tab:additional-gate-transfer-textual}
\small
\begin{tabular}{@{}lllr@{}}
\hline
Type & Method & Backbone & Overall \\
\hline
Textual & Rolling Summary & Qwen3-8B & 26.58 \\
Textual & Mem0 & Qwen3-8B & 30.02 \\
Textual & RAG (BGE-M3) & Qwen3-8B & 30.32 \\
Textual & LightMem & Qwen3-8B & 30.46 \\
Textual & Full Context & Qwen3-8B & 31.10 \\
Parametric & RPMem (PERMA $\to$ PersonaMem-v2, zero-shot) & Qwen3-8B & \textbf{36.25} \\
\hline
\end{tabular}
\end{table}

Under the identical Qwen3-8B backbone, zero-shot RPMem reaches 36.25\%, exceeding Full Context by 5.15 percentage points, LightMem by 5.79 points, RAG by 5.93 points, and Mem0 by 6.23 points. These results provide a direct comparison with textual memory methods and show that PERMA-trained consolidation gates transfer to PersonaMem-v2 without target-benchmark supervision.
\par\endgroup
\par\endgroup

\FloatBarrier
\section{Full Experimental Configuration and Hyperparameters}
\label{app:experimental-configuration}

\subsection{Evaluation Scope and Benchmark Protocols}
\label{app:benchmark-protocols}

\paragraph{\permabench.}
\permabench{} contains ten simulated users with temporally ordered
interaction histories.  Preferences are introduced, extended,
or revised by events in these sessions, and each evaluation query presents
eight candidate answers labeled A through H.  The benchmark crosses two
history structures, single-domain (SD) and multi-domain (MD), with clean and
noisy history realizations to form four core settings. Style SD and Style MD
rewrite the histories in user-specific linguistic styles, while Style-Long SD
adds realistic long-dialogue distractors to the rewritten single-domain
history.  Core Avg. is the unweighted mean of the four core settings.
The seven-setting average is used only in analyses that explicitly report all
seven settings.  Task-adapted methods
use ten-fold leave-one-user-out evaluation: every fold trains on nine users and
reports predictions for the remaining user, after which accuracy is averaged
over the ten held-out users.

\paragraph{\personamembench.}
We use the frozen 32K-Text release, comprising 18,527 training examples, 2,059
validation examples, and 5,000 independent benchmark questions.  Each example
associates a long user history with a multiple-choice query.  The annotations
identify whether the queried information concerns the focal user or another
person and whether the relevant state is current, updated, or marked for
forgetting.  Overall is the official aggregate accuracy.  Self restricts
evaluation to information about the focal user, and Current restricts it to
the currently valid state.  These two slices accompany Overall in the compact
main table because they directly expose the user-specific state maintained
across the history.  The complete ownership and update-status breakdown is
retained in the benchmark result artifacts.

\paragraph{\prefevalbench.}
The evaluation uses the official split of 16 training topics and four held-out
topics.  It covers explicit statements, implicit choice-based evidence, and
implicit persona-driven evidence, each evaluated after 10, 70, and 300 turns
of intervening dialogue.  For each history length, the main table reports the
unweighted mean accuracy over the three evidence forms.  Topic-disjoint
evaluation measures whether the downstream consolidation policy transfers to
semantic domains absent from its training partition.

\subsection{Baseline Implementations}
\label{app:baseline-implementations}

\paragraph{Reference conditions.}
No Context presents the query and candidate answers without historical
information. Gold State converts the state annotation supplied by a
benchmark into structured text and places it before the query.  The latter
condition measures answer selection given direct access to the annotated
state; its result also reflects the coverage and granularity of the benchmark
annotation.

\paragraph{Textual-memory methods.}
Full Context concatenates all visible sessions in chronological order and
places the resulting history before the query.  RAG divides the same visible
history into retrievable units, embeds them with M3-Embedding (BGE-M3), and
injects the ten units with the highest dense-retrieval scores.  On
\permabench{}, units are non-overlapping groups of two consecutive messages
from the chronologically flattened history, including role labels. Retrieved
units are presented in descending relevance order; embedding inputs are
truncated at 8,192 tokens. Rolling Summary
updates a 1,024-token summary after every session by combining the previous
summary with the incoming interaction.  \tbd{The Mem0 baseline uses \texttt{mem0ai} version 2.0.12 (July 2026), including its updated single-pass memory extraction and enabled hybrid semantic/BM25/entity retrieval pipeline. Qwen3-8B is used for memory extraction, with BGE-M3 providing semantic retrieval embeddings. The LightMem baseline uses the April 2026 implementation snapshot and its hierarchical semantic compression and retrieval pipeline. Our two-pass LightMem evaluation extends this implementation with Qwen3-8B-based query refinement and reciprocal rank fusion, as detailed in Appendix~\ref{sec:additional-lightmem-query}.}  Every method receives an
identical query and candidate-answer interface after constructing its textual
memory input.

\paragraph{SFT.}
This trained control freezes Qwen3-8B and fits a rank-8 LoRA with scale 32 on
the \texttt{down\_proj} modules.  For each \permabench{} variant and held-out
user, the adapter is trained for five epochs on the raw history sessions of the
other nine users with the standard assistant-response objective.  The resulting
70 adapters follow the same leave-one-user-out boundary as downstream
consolidation.  Training supervision consists of dialogue responses from the
training histories, while evaluation supplies the held-out user's complete
visible history before each query.  The same construction follows the official
training partition on the other main benchmarks.

\paragraph{Metis.}
We load the released Metis-9B memory checkpoint with its native Qwen3.5-9B
backbone.  Historical sessions are written sequentially through the model's
native recurrent memory mechanism, preserving their original order.  The
memory state is reset at the beginning of every independent benchmark task,
and the released checkpoint is evaluated directly through the shared
non-thinking answer-selection interface.  Results for the released 4B, 9B,
and 27B checkpoints are reported in \appref{app:metis-scale}; the main table
uses the strongest overall scale, Metis-9B.

\subsection{Shared Evaluation and Training Protocol}
\label{app:evaluation-protocol}

In the main comparison, \methodname{} and our baseline implementations use
Qwen3-8B as the answering model. We directly evaluate the released Metis-9B
model. Generalization and transfer use the five backbones reported in
\appref{app:model-generalization-results} and \appref{app:backbone-transfer}.
Thinking mode is disabled throughout. Within each
benchmark, methods receive the same chronological history boundary, query,
candidate answers, and output parser.  On \permabench{}, prediction selects the
largest next-token logit among the legal labels A through H.  Accuracy is first
computed for each held-out user and then macro-averaged across the ten folds.
\personamembench{} uses its frozen training, validation, and benchmark splits,
and \prefevalbench{} uses its official topic-disjoint split.

The default \methodname{} run losslessly segments sessions at 4,096 tokens and
overlaps one complete event between adjacent segments.  Its compiler is the
51,885-update checkpoint trained with the forward-KL objective in
\autoref{eq:compilation-fkl}.  Each stored reference position contains the 32
highest-probability context-conditioned tokens and a tail category containing
the remaining probability mass.  Each accepted compilation session contributes
eight training pairs and two disjoint query-level validation pairs.  During
downstream adaptation, the context encoder, Perceiver-based resampler, LoRA decoder, and
answering model remain fixed while the \consolidationgate{} is optimized on the
training partition.  \autoref{tab:phase2-configuration} records the default
\permabench{} gate configuration; compiler architecture, optimization, and
the remaining downstream protocols appear in \appref{app:compiler-configuration}.

\begin{table}[htbp]
    \centering
    \caption{Primary \consolidationphase{} configuration used for \permabench.}
    \label{tab:phase2-configuration}
    \small
    \begin{tabular}{ll}
        \toprule
        Component & Setting \\
        \midrule
        Backbone & Qwen3-8B, non-thinking mode \\
        Data split & 10-fold leave-one-user-out \\
        Compiler & \fixedfkl{} checkpoint 51,885 \\
        Maximum segment length & 4,096 tokens \\
        Adjacent-segment overlap & One event \\
        Trainable module & Consolidation gate only \\
        Trainable parameters & 524,800 \\
        Initial memory & $h_0=0$; $h_1=q_1$ \\
        Gate initialization & Weight 0, bias $-2.0$ \\
        Optimizer & AdamW \\
        Learning rate & $10^{-3}$ \\
        Weight decay & 0.01 \\
        Gradient clipping & 1.0 \\
        Training epochs & 5 \\
        Batch size & 1 \\
        Random seed & 42 \\
        Prediction & Maximum A--H next-token logit \\
        \bottomrule
    \end{tabular}
\end{table}

\subsection{Compiler Architecture and Optimization}
\label{app:compiler-configuration}

\noindent\textbf{Architecture.}\quad
The primary Qwen3-8B compiler uses a frozen ModernBERT-base context encoder.
Its hidden states are mapped to the 36 target-layer positions by uniformly
spaced encoder indices, with repeated indices when the target has more layers
than the encoder.  LoRA is applied to the feed-forward down-projection
(\texttt{down\_proj}) in each target layer, giving $L=36$ and $M=1$.
With $r=8$ and $d=512$, each session memory has shape
$36\times1\times8\times512$.
\autoref{tab:compiler-architecture} specifies the encoder and decoder.

\begin{table}[htbp]
    \centering
    \caption{Primary Qwen3-8B compiler architecture.}
    \label{tab:compiler-architecture}
    \small
    \begin{tabular}{ll}
        \toprule
        Component & Configuration \\
        \midrule
        Frozen models & Qwen3-8B; ModernBERT-base \\
        Adapted layers / module types & $L=36$; $M=1$, \texttt{down\_proj} \\
        Generated factor width / latent width & $r=8$; $d=512$ \\
        Perceiver latent queries & 8 \\
        Perceiver encoder / output blocks & 9 / 1 cross-attention blocks \\
        Attention heads / head width & 8 / 64 \\
        Perceiver feed-forward width & 2,048 \\
        Additional self-attention blocks & 0 \\
        Decoder pre-head & 4 layer-specific residual MLP blocks \\
        Decoder MLP widths & $512\rightarrow2{,}048\rightarrow512$ \\
        Decoder output & Layer-specific linear factor projections \\
        Shared head-bias block & Enabled; width 8 \\
        LoRA scaling / dropout & $\gamma=32$ / 0 \\
        \bottomrule
    \end{tabular}
\end{table}

Each decoder residual block applies layer normalization, a layer-specific
linear expansion, SiLU, a linear contraction, and a second layer
normalization before adding the residual.  After the four blocks, each latent
vector is normalized by its Euclidean norm and projected to the concatenated
input- and output-factor dimensions.  Splitting this projection yields the
two factors, each multiplied by a learned layer- and rank-specific scalar.
These scalars are initialized to one for $A$ and zero for $B$.
For the shared head-bias block, entries of $A_0$ are initialized from a
zero-mean Gaussian with standard deviation $0.2/\sqrt{r d^{\mathrm{in}}}$,
while $B_0$ is initialized to zero. This block is appended once during
adapter assembly, as defined in \appref{app:consolidation-variants}.
The coefficient $\gamma$ directly multiplies the factor product, using the
same scaling convention across assembled factor widths.

\noindent\textbf{Factor regularization.}\quad
For a bounded training session $c$, the implementation first computes the
mean absolute value of each generated factor, adds the $A$ and $B$ terms,
and averages over layers and module types:
\begin{equation}
    \mathcal{R}_1(c)=\frac{1}{LM}\sum_{\ell=1}^{L}\sum_{m=1}^{M}
    \left(
    \frac{\|A_{\ell,m}(c)\|_1}{r d^{\mathrm{in}}_{\ell,m}}+
    \frac{\|B_{\ell,m}(c)\|_1}{r d^{\mathrm{out}}_{\ell,m}}
    \right).
    \label{eq:factor-regularization}
\end{equation}
Here, $\|\cdot\|_1$ denotes the sum of absolute entries, and
$A_{\ell,m}(c),B_{\ell,m}(c)$ are the generated factors for session $c$ at
layer $\ell$ and module type $m$ after their learned scalar multipliers.
The denominators are the respective entry counts, using the dimensions
defined in \autoref{sec:memory-compiler}.
Both factors receive equal weight despite different input and output widths.
The penalty is evaluated before shared-bias concatenation and application of
$\gamma$, and uses $\lambda=0.01$ in \autoref{eq:compilation-fkl}.

\noindent\textbf{Compilation optimization.}\quad
Training updates the Perceiver-based resampler and decoder while freezing both language
models.  Each of the 664,128 training sessions contributes eight pairs;
response-token losses are averaged within pairs and then within sessions.
The recorded global batch contains 64 sessions, obtained from eight workers,
one session per worker, and eight gradient-accumulation steps.  Five passes
therefore give $664{,}128\times5/64=51{,}885$ optimizer updates.
The final completed checkpoint is used for downstream evaluation.
\autoref{tab:compiler-training} records the primary training configuration.

\begin{table}[htbp]
    \centering
    \caption{Primary fixed-reference forward-KL compilation configuration.}
    \label{tab:compiler-training}
    \small
    \begin{tabular}{ll}
        \toprule
        Component & Setting \\
        \midrule
        Training sessions / pairs per session & 664,128 / 8 \\
        Retained reference support & $K=32$, with aggregate tail mass \\
        Maximum session length & 4,096 tokens \\
        Maximum query-response sequence & 640 tokens \\
        Maximum reference-model sequence & 4,864 tokens \\
        Global batch / accumulation steps & 64 sessions / 8 \\
        Workers / sessions per worker & 8 / 1 \\
        Training passes / updates & 5 / 51,885 \\
        Optimizer & AdamW, $\beta_1=0.9$, $\beta_2=0.999$ \\
        Configured learning rate & $4\times10^{-5}$ \\
        Warmup updates & 500 \\
        Weight decay / gradient clipping & 0.01 / 1.0 \\
        Factor-regularization weight & $\lambda=0.01$ \\
        Random seed & 42 \\
        \bottomrule
    \end{tabular}
\end{table}

\noindent\textbf{Downstream optimization.}\quad
The \permabench{} settings appear in \autoref{tab:phase2-configuration}.
For \personamembench{} and \prefevalbench{}, the formal training protocol
uses the shared configuration in \autoref{tab:downstream-training}.
Each training question produces one optimizer update.  \personamembench{}
shuffles personas and then their training questions using the fixed seed;
the chronological order of the history within each question is preserved.
\prefevalbench{} trains one gate jointly across the 16 training topics,
three information forms, and three primary history lengths, and evaluates
on the four held-out topics.  The data construction and split definitions
appear in \appref{app:consolidation-data}.  In all cases, optimization updates
only the consolidation gate; the compiler and answering backbone remain
frozen.  The final-epoch gate is used for evaluation.

\begin{table}[htbp]
    \centering
    \caption{Shared downstream training configuration for \personamembench{} and \prefevalbench{}.}
    \label{tab:downstream-training}
    \small
    \begin{tabular}{ll}
        \toprule
        Setting & Value \\
        \midrule
        Trainable module & Recurrent consolidation gate \\
        Initial memory & $h_0=0$; $h_1=q_1$ \\
        Gate weight / bias initialization & $W_g=0$; $b_g=-2$ \\
        Training objective & Cross-entropy over legal answer labels \\
        Optimizer / learning rate & AdamW / $10^{-3}$ \\
        Weight decay / gradient clipping & 0.01 / 1.0 \\
        Training epochs / random seed & 5 / 42 \\
        Questions per optimizer update & 1 \\
        Evaluation checkpoint & Final epoch \\
        \bottomrule
    \end{tabular}
\end{table}

\subsection{Training and Query-Cost Details}
\label{app:efficiency-details}

\paragraph{Streaming deployment protocol.}
We profile accumulated source-session counts of
$1,2,4,8,16,32,$ and $64$ on eligible \permabench{} Clean-SD histories.
All models and services are resident before timing.  A hot \methodname{} update includes segmentation and encoding of the newly
arrived session, recurrent memory consolidation, and LoRA materialization.
For adapter rank concatenation, we concatenate the accumulated history,
repartition it into chunks of at most 4,096 tokens, and recompile all chunks
at each update. Unlike the session-wise segmentation used in the consolidation
ablation (\appref{app:consolidation-variants}), these chunks can span
session boundaries. The Mem0 timer includes
LLM-based memory extraction, embedding, and vector-index updates, while Full
Context has no separate write operation.  Persistent memory counts per-user data retained for answering queries and
processing subsequent updates, excluding shared model parameters such as
the compiler and consolidation gate. Historical query tokens count the
memory tokens supplied to the answering model.  Growth values are the empirical
ratios between the 64-session and one-session endpoints rather than asymptotic
complexity claims.

\paragraph{General-capability retention.}
We evaluate the unmodified Qwen3-8B backbone (Base LLM, with no user-memory
adapter) and ten user-specific memory states on
MMLU~\citep{hendrycks2021mmlu}, GSM8K~\citep{cobbe2021gsm8k}, and
IFEval~\citep{zhou2023ifeval}.  \autoref{tab:general-retention} reports the
user-level mean and standard deviation for both memory methods.

\begin{table}[htbp]
    \centering
    \caption{General-benchmark performance (\%) of the base model and memory-adapted models.}
    \label{tab:general-retention}
    \small
    \setlength{\tabcolsep}{5.0pt}
    \begin{tabular}{lccc}
        \toprule
        Method & MMLU (5-shot) & GSM8K (5-shot) & IFEval (0-shot) \\
        \midrule
        Base LLM & 71.96 & 75.21 & 81.70 \\
        Latest Session & $69.85\pm0.51$ & $73.15\pm1.63$ & $77.38\pm1.20$ \\
        \methodname{} (ours) & $70.13\pm0.41$ & $73.78\pm1.53$ & $77.89\pm1.42$ \\
        \bottomrule
    \end{tabular}
\end{table}

\paragraph{Query cost on \personamembench.}
The deployment measurements in the main text are complemented by an
independent evaluation over the 5,000-question \personamembench{} test set.
\autoref{tab:personamem-query-cost} reports the mean input length, language-model
forward time, and parametric state associated with each method.

\begin{table}[htbp]
    \centering
    \caption{Mean query cost on the \personamembench{} test set.}
    \label{tab:personamem-query-cost}
    \small
    \setlength{\tabcolsep}{5.0pt}
    \begin{tabular}{lrrl}
        \toprule
        Method & Input tokens & Forward time (s) & Parametric state \\
        \midrule
        Full Context & 33,363.71 & 4.287 & -- \\
        RAG (BGE-M3) & 2,757.58 & 0.253 & -- \\
        Latest Session & 337.49 & 0.056 & 36 MiB, rank 16 \\
        Adapter rank concatenation & 337.49 & 0.056 & 195.9 MiB, rank 87.1 mean \\
        \textbf{\methodname} & \textbf{337.49} & \textbf{0.055} & \textbf{36 MiB, rank 16} \\
        \bottomrule
    \end{tabular}
\end{table}

The rank reported in \autoref{tab:personamem-query-cost} denotes the assembled
factor width: eight generated components plus the eight-component shared
head-bias block for \methodname{} and Latest Session
(\appref{app:consolidation-variants}), not the numerical rank of the update matrix.
With an approximately 32K-token user history, \methodname{} reduces the mean
forward time from 4.287 seconds under Full Context to 0.055 seconds while
preserving a fixed-rank 36 MiB state.  Rank concatenation receives the same
query tokens but expands the adapter state with the number of accumulated
sessions.
At the 64-session endpoint of the deployment experiment,
\methodname{} incurs 0.575 GiB of peak incremental GPU memory, compared with
2.826 GiB for adapter rank concatenation; this measurement is reported here
because peak allocation is not available for the textual-memory pipelines.

\paragraph{Consolidation adaptation cost.}
For a representative Clean-SD fold with 630 training examples and 3,150
updates, the 524,800-parameter \consolidationgate{} trains in 13.0 minutes on a single GPU
and produces a 2.004 MiB checkpoint.
The peak GPU-memory increase over the full training run is 12.173 GiB,
measured relative to the resident-model baseline before gate training.
This increase primarily reflects activations needed to backpropagate through
the frozen decoder and language model rather than the gate parameters.

\begin{table}[htbp]
    \centering
    \caption{Consolidation-stage adaptation cost on a representative
    \permabench{} fold.}
    \label{tab:consolidation-training-cost}
    \small
    \begin{tabular}{lr}
        \toprule
        Measurement & Value \\
        \midrule
        Trainable parameters & 524,800 \\
        Training examples & 630 \\
        Parameter updates & 3,150 (5 epochs) \\
        Training time & 13.0 min \\
        Peak incremental memory & 12.173 GiB \\
        Saved checkpoint size & 2.004 MiB \\
        \bottomrule
    \end{tabular}
\end{table}

To measure how adaptation scales with the memory sequence, we select ten
Type-3 tasks containing at least 64 cached session representations, truncate
each sequence to progressively longer prefixes, and repeat every measurement
three times.  Peak $\Delta$ reports the mean per-step peak memory increase above a
warmed-up baseline with the model, gate, optimizer, and query tensors
already resident.  As shown in \autoref{tab:consolidation-length-cost}, increasing
the sequence from one to 64 representations raises total step time by
$1.39\times$ and peak incremental memory by 3.6\%.

\begin{table}[htbp]
    \centering
    \caption{Consolidation optimization cost as the number of cached session
    representations increases.}
    \label{tab:consolidation-length-cost}
    \small
    \setlength{\tabcolsep}{5.0pt}
    \begin{tabular}{rrrrr}
        \toprule
        Sessions & Load (ms) & Compute (ms) & Total (ms) & Peak $\Delta$ (GiB) \\
        \midrule
        1  & 0.8  & 159.0 & 159.8 & 5.059 \\
        2  & 1.4  & 161.1 & 162.5 & 5.062 \\
        4  & 2.6  & 161.6 & 164.3 & 5.069 \\
        8  & 4.7  & 161.2 & 165.9 & 5.081 \\
        16 & 8.3  & 164.1 & 172.4 & 5.107 \\
        32 & 17.3 & 169.6 & 186.9 & 5.153 \\
        64 & 43.1 & 179.6 & 222.7 & 5.242 \\
        \bottomrule
    \end{tabular}
\end{table}

Of the additional 62.9 ms at 64 sessions, 42.3 ms is spent loading cached
representations and 20.6 ms in GPU computation.  The measured scaling therefore
localizes the principal growth term to cache I/O, while the recurrent
consolidation computation and activation memory remain comparatively stable.

\paragraph{Compilation optimization cost.}
\autoref{tab:compilation-training-cost} reports measured step times and
estimated allocated GPU-hours for the objectives evaluated in
\autoref{tab:objective-ablation}.
A fixed-reference full-vocabulary reverse-KL run (\fixedrkl{}) is included
as an additional runtime comparison.
The three 8-GPU configurations have nearly identical step times and
an estimated cost of 338--339 GPU-hours for 51,885 updates.
Each of the four 24-GPU runs, including \fixedrkl{}, performs 10,377 SFT
warm-start updates followed by 41,508 updates with its respective objective.
The total estimate uses separately measured preflight step times for the
two phases:
\[
    \text{GPU-hours}
    =\frac{24}{3600}\left(10{,}377\,t_{\mathrm{SFT}}
    +41{,}508\,t_{\mathrm{objective}}\right).
\]
Here, both step times are in seconds; only $t_{\mathrm{objective}}$ is
displayed for these runs in the table.

\begin{table}[htbp]
    \centering
    \caption{Compilation-stage optimization cost by training objective.}
    \label{tab:compilation-training-cost}
    \small
    \setlength{\tabcolsep}{4.5pt}
    \begin{tabular}{lllrrr}
        \toprule
        Objective & Prefix & Loss & Step (s) & Allocated GPUs & Est. GPU-hours \\
        \midrule
        Top-$K$ CE & Reference & Top-$K$ CE & 2.943 & 8 & 339 \\
        \textbf{\fixedfkl{}} & Reference & FKL & \textbf{2.932} & 8 & \textbf{338} \\
        \compilersft{} & Reference & CE & 2.940 & 8 & 339 \\
        \sampledrkl{} (K3) & Student & RKL (K3) & 13.061 & 24 & 3,849 \\
        \fixedrkl{} & Reference & RKL & 9.585 & 24 & 2,939 \\
        \sampledfkl{} & Student & FKL & 18.202 & 24 & 5,284 \\
        \sampledrkl{} (top-$K$) & Student & RKL & 15.944 & 24 & 4,627 \\
        \bottomrule
    \end{tabular}
\end{table}

The 24-GPU runs reserve 16 actor and eight teacher GPUs throughout training.
Objective-stage timing includes online teacher scoring and, for sampled
variants, student generation. Estimates exclude non-update overhead and
separate offline reference preparation.
\fixedfkl{} combines the highest downstream accuracy in
\autoref{tab:objective-ablation} with the lowest estimated allocated training
cost among the evaluated distribution-matching configurations.

\subsection{Complete Historical-Context Shift Results}
\label{app:perma-style-results}

\autoref{tab:perma-style-full} reports the complete method matrix for the three
historical-context shift settings summarized in \autoref{sec:main-results}.
Style SD and Style MD rewrite the visible histories in user-specific linguistic
styles.  Style-Long SD additionally inserts realistic long-dialogue distractors
into the rewritten single-domain history.  Avg. is the unweighted mean of the
three settings.

\begin{table}[htbp]
    \centering
    \caption{Complete \permabench{} historical-context shift results (accuracy, \%).}
    \label{tab:perma-style-full}
    \small
    \setlength{\tabcolsep}{5.2pt}
    \begin{tabular}{lrrrr}
        \toprule
        Method & Style SD & Style-Long SD & Style MD & Avg. \\
        \midrule
        Full Context & 76.40 & 75.83 & 71.35 & 74.53 \\
        Gold State & 73.50 & 73.50 & 63.80 & 70.27 \\
        RAG (BGE-M3) & 65.53 & 65.53 & 46.47 & 59.18 \\
        Rolling Summary & 39.63 & 39.93 & 26.66 & 35.41 \\
        Mem0 & 58.73 & 59.47 & 40.35 & 52.85 \\
        LightMem & 55.13 & 56.17 & 39.34 & 50.21 \\
        SFT & 76.73 & 77.40 & 56.68 & 70.27 \\
        Metis-9B & \underline{79.67} & \underline{79.97} &
        \underline{79.87} & \underline{79.84} \\
        \rowcolor{black!6}
        \textbf{\methodname{} (ours)} & \textbf{91.37} & \textbf{90.97} &
        \textbf{81.33} & \textbf{87.89} \\
        \bottomrule
    \end{tabular}
\end{table}

\methodname{} leads all three context-shift variants and reaches an average
accuracy of 87.89\%, exceeding Metis-9B and Full Context by 8.05 and 13.36
percentage points (pp), respectively.
Relative to the corresponding clean settings, its accuracy changes by
$+0.80$ pp on Style SD, $+0.40$ pp on Style-Long SD, and $+1.66$ pp on
Style MD.
Style-Long SD jointly varies linguistic style and history length, so its result
measures stability under the combined shift.
Across all seven \permabench{} variants, \methodname{} reaches
\permasevenacc{}, demonstrating that the parametric memory remains reliable as
the style and length of the historical context change.

\subsection{Complete Model-Generalization Results}
\label{app:model-generalization-results}

\autoref{tab:model-generalization-full} reports the aggregate values underlying
\autoref{fig:model-results}.  SD Avg. averages the four single-domain variants,
MD Avg. averages the three multi-domain variants, and Overall Avg. averages all
seven \permabench{} variants.  Within each backbone, all methods use the same
history boundary, non-thinking answer-selection protocol, and held-out-user
folds. No Context results are aggregated from independent evaluations of each variant.

\begin{table}[htbp]
    \centering
    \caption{Complete aggregate results underlying the cross-model
    generalization comparison (accuracy, \%).}
    \label{tab:model-generalization-full}
    \small
    \setlength{\tabcolsep}{5.0pt}
    \begin{tabular}{llrrr}
        \toprule
        Backbone & Method & SD Avg. & MD Avg. & Overall Avg. \\
        \midrule
        \multirow{3}{*}{Qwen3-4B}
        & No Context & 65.70 & 60.97 & 63.67 \\
        & Full Context & 79.82 & 73.39 & 77.06 \\
        & \textbf{\methodname{} (ours)} & \textbf{85.98} & \textbf{79.04} & \textbf{83.01} \\
        \addlinespace
        \multirow{3}{*}{Qwen3-8B}
        & No Context & 66.73 & 54.60 & 61.53 \\
        & Full Context & 75.43 & 70.67 & 73.39 \\
        & \textbf{\methodname{} (ours)} & \textbf{91.03} & \textbf{80.54} & \textbf{86.53} \\
        \addlinespace
        \multirow{3}{*}{Ministral-3-8B}
        & No Context & 69.57 & 74.47 & 71.67 \\
        & Full Context & 77.05 & 77.56 & 77.27 \\
        & \textbf{\methodname{} (ours)} & \textbf{89.48} & \textbf{83.00} & \textbf{86.70} \\
        \addlinespace
        \multirow{3}{*}{Qwen3.5-9B}
        & No Context & 29.24 & 35.63 & 31.98 \\
        & Full Context & 66.67 & 84.34 & 74.25 \\
        & \textbf{\methodname{} (ours)} & \textbf{96.21} & \textbf{92.59} & \textbf{94.66} \\
        \addlinespace
        \multirow{3}{*}{Qwen3.5-35B-A3B}
        & No Context & 76.00 & 84.11 & 79.48 \\
        & Full Context & 57.72 & 74.38 & 64.86 \\
        & \textbf{\methodname{} (ours)} & \textbf{93.65} & \textbf{94.96} & \textbf{94.21} \\
        \bottomrule
    \end{tabular}
\end{table}

\FloatBarrier
\begingroup
\subsection{Iterative Retrieval and Query-Time Efficiency}
\label{sec:additional-lightmem-query}

We compare RPMem with single-pass and two-pass LightMem on the 540 questions in the PrefEval 300-turn setting. The two retrieval configurations use the same task and evaluation questions. Table~\ref{tab:additional-lightmem-query} reports answer accuracy, mean query latency, and language-model tokens processed per query.

\begingroup
\noindent\textbf{Two-pass LightMem implementation.}\quad
We first retrieve the top 10 memories from LightMem using the original question. We then provide Qwen3-8B with the question and these retrieved memories, prompting it to generate one short follow-up search query targeting potentially missing constraints, exceptions, or later changes in user preferences. Using this generated query, we retrieve another top 10 memories from the same memory index. We merge the two ranked lists using reciprocal rank fusion (RRF, $k = 60$), deduplicate by memory ID, and retain only the 10 highest-ranked memories for the final reader to answer the original question.

Thus, the two-pass setting adds an intermediate query-generation and retrieval step while keeping the final reader's memory budget fixed at 10 retrieved items, identical to the single-pass setting.
\par\endgroup

\begin{table}[htbp]
\centering
\caption{Query-time accuracy and efficiency on PrefEval with 300 intervening turns.}
\label{tab:additional-lightmem-query}
\small
\begin{tabular}{lrrr}
\hline
Method & Accuracy (\%) & Latency (ms) & LM tokens/query \\
\hline
RPMem & 75.19 & \textbf{146.24} & \textbf{112.39} \\
LightMem, single-pass & 74.81 & 326.88 & 566.35 \\
LightMem, two-pass & \textbf{75.74} & 1518.25 & 1116.85 \\
\hline
\end{tabular}
\end{table}

The second retrieval pass increases LightMem accuracy from 74.81\% to 75.74\%. RPMem achieves 75.19\%, within 0.55 percentage points of two-pass LightMem, with 10.38 times faster query processing and 9.94 times fewer LM tokens. Compared with single-pass LightMem, RPMem reduces query latency by 55.26\% while obtaining slightly higher accuracy. Explicit retrieval offers direct access to source evidence, while the consolidated parametric memory avoids repeated processing of historical memory text at each query.

\subsection{Amortized Deployment Cost under Backbone Replacement}
\label{sec:additional-deployment-cost}

We estimate cumulative memory-operation and backbone-adaptation cost over a one-year deployment. The model-level Qwen3.5-9B decoder adaptation takes 30.92 hours on eight GPUs, corresponding to 247.36 GPU-hours per newly supported backbone (Table G.3). This decoder is trained once per target backbone and shared across users. For online operation we use RPMem query latency of 0.14624 s and single-pass LightMem query latency of 0.32688 s from the PrefEval 300-turn comparison. The measured RPMem write time is 0.043 s/session in the 64-session streaming profile. LightMem ingests a 301-session PrefEval history in 177.63 s, corresponding to an amortized 0.590 s/session. The latter is an average over whole-history ingestion, not an independently timed incremental write. We assume one session write per user per day and convert online timings to single-GPU-equivalent hours; decoder training uses the measured allocation. Annual, quarterly, monthly, and weekly replacement correspond to 1, 4, 12, and 52 decoder adaptations per year.

\begin{table}[htbp]
\centering
\caption{Break-even memory queries per user per day across backbone replacement frequencies and user populations.}
\label{tab:additional-cost-breakeven}
\small
\begin{tabular}{lrrrr}
\hline
Replacement frequency & 1K users & 10K users & 100K users & 1M users \\
\hline
Annually & 10.48 & 0.00 & 0.00 & 0.00 \\
Quarterly & 51.00 & 2.37 & 0.00 & 0.00 \\
Monthly & 159.04 & 13.18 & 0.00 & 0.00 \\
Weekly & 699.28 & 67.20 & \textbf{4.00} & \textbf{0.00} \\
\hline
\end{tabular}
\end{table}

A break-even value of zero indicates that the estimated writing-cost savings alone cover the shared decoder-adaptation cost. For example, with weekly backbone replacements, the threshold is 67.20 queries per user per day at 10K users and 4.00 at 100K users.

\begin{table}[H]
\centering
\caption{Estimated annual adaptation, writing, and querying cost for 100K users, ten queries and one session write per user per day (GPU-equivalent hours).}
\label{tab:additional-cost-annual}
\small
\begin{tabular}{lrrrr}
\hline
Replacement frequency & RPMem total & LightMem total & Net saving & Reduction (\%) \\
\hline
Annually & 15510.44 & 39123.94 & 23613.50 & 60.36 \\
Quarterly & 16252.52 & 39123.94 & 22871.42 & 58.46 \\
Monthly & 18231.40 & 39123.94 & 20892.54 & \textbf{53.40} \\
Weekly & 28125.80 & 39123.94 & 10998.14 & \textbf{28.11} \\
\hline
\end{tabular}
\end{table}

At this workload, the estimated cumulative computation cost is reduced by 53.40\% with monthly and 28.11\% with weekly backbone replacement, including decoder training and online memory operations. A shared decoder-adaptation investment is amortized across users and subsequent queries.
\par\endgroup

\FloatBarrier
\section{Memory Dynamics Analysis Details}
\label{app:memory-dynamics-details}

\autoref{tab:lifecycle-results} reports the functional lifecycle comparison
that motivates the update-dynamics analysis in the main text.  T1 precedes the
target memory event, T2 immediately follows it, and T3 follows subsequent
intervening sessions.  The table reports the memory-bearing T2 and T3
checkpoints averaged across all seven \permabench{} variants.

\begin{table}[htbp]
    \centering
    \caption{Memory-bearing accuracy (\%) across the \permabench{} lifecycle.}
    \label{tab:lifecycle-results}
    \small
    \setlength{\tabcolsep}{5.0pt}
    \begin{tabular}{lccc}
        \toprule
        Method & In-Time (T2) & Post-Intervention (T3) & T3--T2 (pp) \\
        \midrule
        Full Context & 76.3 & 77.2 & $+0.9$ \\
        Latent averaging & 46.5 & 45.6 & $-0.9$ \\
        \textbf{\methodname{} (ours)} & \textbf{86.7} & \textbf{86.5} & $-0.2$ \\
        \bottomrule
    \end{tabular}
\end{table}

\subsection{Source Decomposition of the Recurrent State}
\label{app:source-decomposition}

With the direct initialization $h_1=q_1$, unrolling the recurrent update in
\eqref{eq:consolidation-update} for $t\geq2$ assigns an explicit
coordinate-wise weight to every source session:
\begin{align}
    h_T
    &=\sum_{i=1}^{T}w_{i,T}\odot q_i,
    \label{eq:cmp-expansion}\\
    w_{1,T}
    &=\prod_{k=2}^{T}z_k,
    \label{eq:cmp-first-source-weight}\\
    w_{i,T}
    &=(1-z_i)\odot\prod_{k=i+1}^{T}z_k,\qquad 2\leq i\leq T.
    \label{eq:cmp-source-weight}
\end{align}
Here, $i$ indexes the source session, $k$ indexes a subsequent update, and
$w_{i,T}$ has the same shape as $q_i$.  Its entries are the coefficients of
that source in the accumulated memory after step $T$.  All products are
element-wise, and an empty product is the all-ones tensor.  Thus $w_{1,1}=1$,
while $w_{i,i}=1-z_i$ for $i\geq2$.  The coefficients are nonnegative and sum
to one at every coordinate.  They describe the source decomposition along
the observed gate trajectory; perturbing a session can also change later
gates and therefore requires a new forward replay.

\subsection{Trajectory Replay and Statistical Protocol}
\label{app:dynamics-protocol}

The analysis uses the Qwen3-8B compiler trained with fixed-reference
forward KL. For each of the ten held-out \permabench{} users, trajectories
are replayed with the consolidation module trained on the other nine users
for the corresponding benchmark variant. Clean SD and Noisy SD each contain
423 task-conditioned trajectories and 34,218 session occurrences, giving
846 trajectories and 68,436 session occurrences in total. These counts
include shared history appearing in different task-conditioned trajectories;
the independent evaluation units are the ten users. Each semantic session
in this replay occupies one compiler segment.

For matched comparisons, we first average the paired differences within
each user and then average the ten user estimates with equal weight.
The 95\% confidence intervals are the 2.5th and 97.5th percentiles of
10,000 bootstrap means obtained by resampling users with replacement.
Matching remains fixed during resampling. These intervals summarize
variation across held-out users conditional on the fitted fold models.

\noindent\textbf{Reading the main dynamics figure.}\quad
In \autoref{fig:memory-dynamics}(a--b), each colored point is one held-out
user's paired-condition mean, and the hollow diamond averages the ten users.
The diagonal marks equal write magnitude. In (c--d), violins show the
matched-pair distributions, overlaid boxes span the interquartile range with
median lines and 5th--95th percentile whiskers, and colored points show user
means. Panel (e) reports each user's paired difference and its across-user
mean; horizontal error bars give the user-bootstrap 95\% confidence interval.
The differences are Same minus Cross for write-pattern similarity and Cross
minus Same for one-step survival, so positive values indicate stronger
same-domain alignment and revision, respectively. In (f), the shaded bands
give user-bootstrap 95\% confidence intervals for the mean survival curves.

Event-role matching is performed without replacement within the same user,
trajectory, and segment count. A minimum-total-cost assignment uses the
sum of absolute differences in normalized session position and log token
count, each scaled by its within-stratum interquartile range. A zero
interquartile range is replaced by one. We report the resulting comparisons
as associations along the observed histories and assess residual covariate
balance using absolute standardized mean differences.

\subsection{Session-Level Write Statistics}

For a semantic session $e$ containing one or more compiler segments, let
$Z_e$ denote the coordinate-wise retention of memory present before the event.
For events following initialization, $Z_e=\prod_{k\in e}z_k$, where $k$
indexes the event's compiler segments.  The event containing the first
segment has $Z_e=0$ because that segment directly initializes memory.
We summarize the resulting update by the scalar
write magnitude
\begin{equation}
    w_e=1-\operatorname{mean}(Z_e).
    \label{eq:session-write-magnitude}
\end{equation}
The mean is taken over all layer, module, rank, and latent-feature coordinates
of $Z_e$. Retention and write statistics are computed over fusion steps $t\geq2$,
following the direct initialization $h_1=q_1$. Within each trajectory, the
write range is the maximum minus the minimum write magnitude over these
fusion steps; the reported trace range averages these ranges across
trajectories. The untrained gate assigns $z=\sigma(-2)=0.119$ to every
coordinate of each gated update. Downstream training raises the
mean retention to 0.670 and produces an average within-trajectory range of
0.141 in write magnitude, as shown in \autoref{tab:gate-statistics}.

\begin{table}[htbp]
    \centering
    \caption{Consolidation statistics on \permabench{} Clean SD for fusion steps $t\geq2$.}
    \label{tab:gate-statistics}
    \small
    \setlength{\tabcolsep}{4.0pt}
    \begin{tabular}{lcccc}
        \toprule
        Consolidation & Retain mean & Retain P5/P50/P95 & Write mean & Trace range \\
        \midrule
        Untrained gate & 0.119 & 0.119/0.119/0.119 & 0.881 & 0.000 \\
        Learned gate & 0.670 & 0.636/0.672/0.698 & 0.330 & 0.141 \\
        \bottomrule
    \end{tabular}
\end{table}

Domain-emergence and supplement events correspond to \permabench{}'s
preference-emergence and preference-supplement annotations, respectively.
The semantic-role comparison uses 810 matched emergence--supplement pairs.
The mean write difference is $+0.056$ with a user-level
bootstrap 95\% confidence interval of $[0.039,0.073]$. After matching, the
absolute standardized differences are 0.055 for log token count and 0.962
for normalized position. The association therefore reflects semantic role
together with its temporal placement in the benchmark.

The Clean/Noisy comparison aligns sessions by user, task identifier and
type, session index, date, event type, domain, and trajectory role. All
34,218 session occurrences align across the two variants; the reported
perturbation comparison selects the 18,234 pairs whose conversation text
differs. Noise changes text within the aligned histories while preserving
their event structure. Each variant uses its corresponding trained gate,
so this comparison measures the behavior of the complete trained recurrence
under the two benchmark conditions. The mean Noisy-minus-Clean difference
is $+0.011$, and the mean absolute per-session drift is 4.8\% of the Clean
write magnitude.

\subsection{Domain Updates and Long-Term Survival}

For the domain analysis, we flatten, center, and normalize the coordinate-wise
write pattern $1-Z_e$ and compute cosine similarity for matched same-domain
and cross-domain event pairs.  One-step survival is the ratio between a
historical source coefficient immediately after and before a new update.
Same-domain and cross-domain write-pattern similarities are 0.872 and 0.784,
respectively; the corresponding one-step survival values are 0.458 and 0.818.
Write-pattern pairs are matched within user and trajectory, with equal
source and target segment counts. Matching covariates include both event
positions, their temporal separation, and both log token counts. Transition
comparisons similarly match source age, incoming-event position, and both
log token counts. These comparisons characterize domain-associated update
patterns in the observed trajectories.

At recurrence-step resolution, source $i$'s survival after $a$ subsequent
updates is
\begin{equation}
    L_i(a)=\frac{w_{i,i+a}}{w_{i,i}},
    \label{eq:source-survival}
\end{equation}
where $a$ is a nonnegative integer and division is element-wise.
The source coefficients are defined by
\autoref{eq:cmp-first-source-weight} for $i=1$ and
\autoref{eq:cmp-source-weight} for $i\geq2$.
For either case, $L_i(a)=\prod_{k=i+1}^{i+a}z_k$, with $L_i(0)=1$.
For multi-segment semantic sessions, subsequent segment retentions are
composed over the corresponding event boundaries.
For the scalar survival curves, we divide the coordinate mean of the
remaining source coefficient by its coordinate mean at insertion. This
weights coordinate-wise survival by the source's initial write coefficients.
At each session age, eligible sources are those whose recorded trajectory
extends to that age. Their scalar values are averaged within user and then
equally across eligible users. All five checkpoints below include ten users.
The emergence-source count is 810 at each checkpoint; supplement-source
counts are 2,871 at ages 1, 5, and 10, 2,502 at age 40, and 1,782 at age 60.
\autoref{tab:source-survival-values}
reports the numerical checkpoints underlying the survival curves in the main
text.

\begin{table}[H]
    \centering
    \caption{Mean source survival after subsequent session updates.}
    \label{tab:source-survival-values}
    \small
    \setlength{\tabcolsep}{6.0pt}
    \begin{tabular}{lrrrrr}
        \toprule
        Source role & 1 & 5 & 10 & 40 & 60 \\
        \midrule
        Domain emergence & 0.294 & 0.121 & 0.081 & 0.038 & 0.032 \\
        Supplement & 0.225 & 0.058 & 0.028 & 0.008 & 0.005 \\
        \bottomrule
    \end{tabular}
\end{table}

At segment resolution, the untrained gate follows the constant decay
$\sigma(-2)^a\approx0.119^a$ and reaches $2.41\times10^{-5}$ after five
gated updates.  A semantic-session horizon composes all intervening segment
updates.  At the five-session checkpoint, the learned
recurrence retains 0.121 of emergence sources and 0.058 of supplement sources.

\FloatBarrier
\section{Full Cross-Backbone Transfer Results}
\label{app:backbone-transfer}

\subsection{Transfer Protocol and Paired Evaluation}
\label{app:transfer-protocol}

\noindent\textbf{Source and target modules.}\quad
Head Transfer reuses the session encoder from the Qwen3-8B compiler at
update 51,885. Its context encoder $E_\xi$ and Perceiver-based resampler $G_\phi$ remain
fixed while a target-specific decoder $D_{\beta'}$ is trained for the frozen
target backbone $f_{\theta'}$. From Scratch initializes the resampler and
target decoder anew and optimizes both, with the context encoder and target
backbone held fixed.  The two conditions therefore compare reuse of the
learned memory encoding with target-side learning from random initialization.
At backbone replacement, the adapted decoder reads the retained memory
$h_T$ and produces $D_{\beta'}(h_T)$ for the new backbone, following
\autoref{alg:deployment}. The memory shape, session encoder, and learned
consolidation gate remain unchanged; the target decoder accommodates the
new backbone's layer count and module dimensions.

\noindent\textbf{Target-side training.}\quad
Both conditions use the same 664,128 compilation sessions and fixed reference
responses. Reference distributions are prepared with the corresponding
target backbone conditioned on each textual session; the paired runs for
that target share these cached distributions.  Each run completes one pass,
10,377 updates, with global batch size 64 and the same random seed, using the
fixed-reference forward-KL objective.  The source compiler's five-pass
training precedes this target-side comparison and supplies the checkpoint
shared across transfer targets.  The matched budget refers to the one-pass
target training stage.

The target configurations retain eight Perceiver queries, latent width 512,
LoRA rank eight, nine Perceiver encoder blocks, and a four-block decoder
pre-head.  Dense targets adapt the MLP down-projection in each layer.
For Qwen3.5-35B-A3B, the decoder generates factors for the always-active shared
expert's down-projection.  The documented optimizer configuration uses AdamW
with learning rate $4\times10^{-5}$, weight decay 0.01, 500 warmup updates,
gradient clipping at 1.0, and seed 42.  The factor regularization coefficient
is 0.01, and the LoRA scaling coefficient is 32.

\noindent\textbf{Batch construction and runtime alignment.}\quad
Qwen3-4B, Ministral-3-8B, and Qwen3.5-9B use eight single-GPU data-parallel replicas,
each processing one session per microbatch and accumulating eight
microbatches.  The Qwen3.5-35B-A3B protocol distributes each frozen backbone
replica over eight GPUs and runs four replicas.  Each replica accumulates
16 sessions, and trainable-module gradients are averaged across replicas.
Both layouts yield 64 sessions per optimizer update, with the same batch
construction for Head Transfer and From Scratch. Validation is scheduled
every 5,000 updates and at the final epoch boundary; the completed one-pass
checkpoint supplies the downstream compiler.  For Qwen3.5 targets, reference
target preparation and compiler training use matching frozen Transformers
and Tokenizers runtimes.  Launch-time checks compare these runtimes with the
target asset manifest to preserve token and response-position alignment.

\noindent\textbf{Downstream evaluation.}\quad
For the paired compiler comparison, each resulting compiler is frozen before
fitting its consolidation gate with
the same architecture and ten-fold leave-one-user-out \permabench{} protocol.
Gate training uses nine users in each fold, and the remaining user supplies
the reported evaluation examples.  This downstream task-adaptation stage is
separate from training the target decoder on the compilation corpus.
\autoref{tab:transfer-sd-full} and \autoref{tab:transfer-md-full} report the
complete paired From Scratch and Head Transfer results over all seven
\permabench{} variants.
\autoref{fig:model-results} reports the corresponding \methodname{} performance
for each language model.

\begin{figure}[htbp]
    \centering
    \includegraphics[
        width=\linewidth,
        trim=7bp 7bp 7bp 10bp,
        clip
    ]{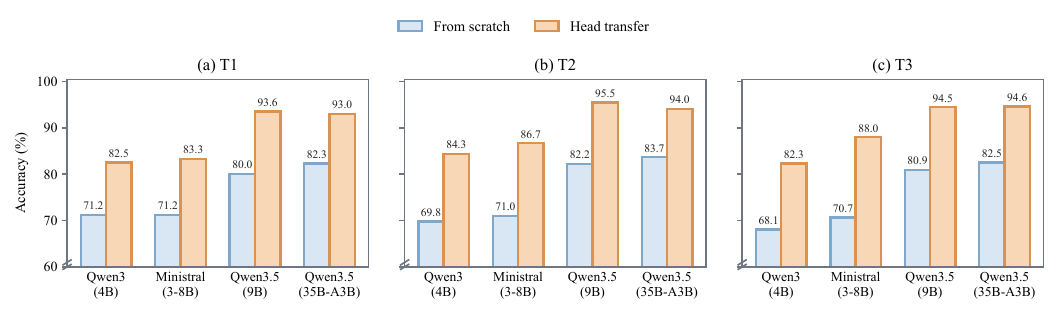}
    \caption{Cross-backbone transfer across the \permabench{} memory
    lifecycle (accuracy, \%; seven-variant average). Vertical axes begin at
    60\%.}
    \label{fig:transfer-lifecycle}
\end{figure}

\begin{table}[htbp]
    \centering
    \caption{Cross-backbone accuracy (\%) on the four single-domain
    \permabench{} variants.}
    \label{tab:transfer-sd-full}
    \small
    \resizebox{\linewidth}{!}{%
    \begin{tabular}{llrrrrr}
        \toprule
        Target backbone & Initialization & Clean SD & Noisy SD & Style SD & Style-Long SD & SD Avg. \\
        \midrule
        \multirow{2}{*}{Qwen3-4B}
          & From scratch & 75.67 & 75.27 & 75.30 & 74.57 & 75.20 \\
          & \textbf{Head transfer} & \textbf{86.20} & \textbf{85.83} & \textbf{85.83} & \textbf{86.07} & \textbf{85.98} \\
        \addlinespace
        \multirow{2}{*}{Ministral-3-8B}
          & From scratch & 76.23 & 76.40 & 75.97 & 76.37 & 76.24 \\
          & \textbf{Head transfer} & \textbf{90.03} & \textbf{89.37} & \textbf{89.27} & \textbf{89.23} & \textbf{89.48} \\
        \addlinespace
        \multirow{2}{*}{Qwen3.5-9B}
          & From scratch & 79.10 & 79.57 & 78.57 & 78.53 & 78.94 \\
          & \textbf{Head transfer} & \textbf{96.50} & \textbf{96.50} & \textbf{96.00} & \textbf{95.83} & \textbf{96.21} \\
        \addlinespace
        \multirow{2}{*}{Qwen3.5-35B-A3B}
          & From scratch & 80.53 & 82.57 & 81.13 & 81.07 & 81.33 \\
          & \textbf{Head transfer} & \textbf{93.53} & \textbf{93.00} & \textbf{94.37} & \textbf{93.70} & \textbf{93.65} \\
        \midrule
        \multirow{2}{*}{Cross-model average}
          & From scratch & 77.88 & 78.45 & 77.74 & 77.63 & 77.93 \\
          & \textbf{Head transfer} & \textbf{91.57} & \textbf{91.18} & \textbf{91.37} & \textbf{91.21} & \textbf{91.33} \\
        \bottomrule
    \end{tabular}}
\end{table}

\begin{table}[htbp]
    \centering
    \caption{Cross-backbone accuracy (\%) on the three multi-domain
    \permabench{} variants and all seven variants.}
    \label{tab:transfer-md-full}
    \small
    \resizebox{\linewidth}{!}{%
    \begin{tabular}{llrrrrr}
        \toprule
        Target backbone & Initialization & Clean MD & Noisy MD & Style MD & MD Avg. & Overall Avg. \\
        \midrule
        \multirow{2}{*}{Qwen3-4B}
          & From scratch & 62.47 & 60.03 & 60.95 & 61.15 & 69.18 \\
          & \textbf{Head transfer} & \textbf{78.89} & \textbf{79.32} & \textbf{78.92} & \textbf{79.04} & \textbf{83.01} \\
        \addlinespace
        \multirow{2}{*}{Ministral-3-8B}
          & From scratch & 63.64 & 63.79 & 63.79 & 63.74 & 70.88 \\
          & \textbf{Head transfer} & \textbf{81.70} & \textbf{84.72} & \textbf{82.59} & \textbf{83.00} & \textbf{86.70} \\
        \addlinespace
        \multirow{2}{*}{Qwen3.5-9B}
          & From scratch & 83.33 & 84.63 & 83.97 & 83.98 & 81.10 \\
          & \textbf{Head transfer} & \textbf{92.52} & \textbf{92.68} & \textbf{92.57} & \textbf{92.59} & \textbf{94.66} \\
        \addlinespace
        \multirow{2}{*}{Qwen3.5-35B-A3B}
          & From scratch & 84.40 & 86.00 & 83.22 & 84.54 & 82.70 \\
          & \textbf{Head transfer} & \textbf{95.17} & \textbf{94.97} & \textbf{94.73} & \textbf{94.96} & \textbf{94.21} \\
        \midrule
        \multirow{2}{*}{Cross-model average}
          & From scratch & 73.46 & 73.61 & 72.98 & 73.35 & 75.97 \\
          & \textbf{Head transfer} & \textbf{87.07} & \textbf{87.92} & \textbf{87.20} & \textbf{87.40} & \textbf{89.64} \\
        \bottomrule
    \end{tabular}}
\end{table}

\subsection{Target-Backbone Adaptation Cost}
\label{app:transfer-cost-scope}

\autoref{tab:transfer-cost} reports the target-side compilation stage defined
above.  Trainable parameters count the optimized decoder weights in Head
Transfer and the resampler plus decoder weights in From Scratch. The
decoder is a shared adapter-generating network; the per-user memory and its
generated LoRA factors are measured separately in
\appref{app:efficiency-details}.
Time is the recorded wall-clock duration of the one-pass target training run.
The first three targets use eight GPUs on one node, while Qwen3.5-35B-A3B
uses 32 GPUs across four nodes.  Timing comparisons are paired within each
target and hardware allocation.  Validation loss is the final checkpoint's
fixed-reference forward-KL validation metric on the compilation data.

\begin{table}[htbp]
    \centering
    \caption{One-pass target-backbone compilation cost.}
    \label{tab:transfer-cost}
    \small
    \setlength{\tabcolsep}{4.0pt}
    \begin{tabular}{llrrrr}
        \toprule
        Target backbone & Initialization & Trainable params & GPUs & Time (h) & Val. loss \\
        \midrule
        \multirow{2}{*}{Qwen3-4B}
          & From scratch & 580.67M & 8 & 7.64 & 0.7568 \\
          & \textbf{Head transfer} & \textbf{532.40M} & 8 & \textbf{6.74} & \textbf{0.6759} \\
        \addlinespace
        \multirow{2}{*}{Ministral-3-8B}
          & From scratch & 659.72M & 8 & 8.16 & 0.6827 \\
          & \textbf{Head transfer} & \textbf{611.45M} & 8 & \textbf{7.46} & \textbf{0.5944} \\
        \addlinespace
        \multirow{2}{*}{Qwen3.5-9B}
          & From scratch & 589.67M & 8 & 32.27 & 0.6486 \\
          & \textbf{Head transfer} & \textbf{541.40M} & 8 & \textbf{30.92} & \textbf{0.5815} \\
        \addlinespace
        \multirow{2}{*}{Qwen3.5-35B-A3B}
          & From scratch & 437.48M & 32 & 41.66 & 0.5713 \\
          & \textbf{Head transfer} & \textbf{389.21M} & 32 & \textbf{38.42} & \textbf{0.5121} \\
        \bottomrule
    \end{tabular}
\end{table}

Freezing the transferred resampler reduces target-side trainable parameters
by 7.32\% to 11.03\% across the four backbones.
Adaptation time decreases by 4.17\% to 11.74\%, and the final \fixedfkl{}
validation loss is lower for every target.  Together with the paired
downstream results, these measurements show that the retained encoder
improves target-side learning under the same target-data exposure.

The cost table accounts for target-side training after the source compiler
and reference targets have been prepared.  Source compilation is the shared
upstream training investment described in \appref{app:compiler-configuration}.
Downstream gate fitting has its own per-fold cost, reported in
\appref{app:efficiency-details}, which also defines the forward-only write
and query measurements.  Thus the reported training durations and deployment
latencies refer to distinct stages of the memory lifecycle.

\FloatBarrier
\begingroup
\subsection{Continued Memory Updates after Backbone Replacement}
\label{sec:additional-backbone-continuation}

We evaluate the full memory lifecycle following a serving-backbone replacement. On PERMA Noisy SD we use ten leave-one-user-out folds and 695 questions with multi-session histories. We test both Qwen3-4B to Qwen3-8B and Qwen3-8B to Qwen3-4B. For each history, the first half of the sessions is encoded and consolidated using the source-trained gate $G_A$. At the midpoint we preserve the accumulated memory state, session encoder, and gate $G_A$, and replace only the serving backbone and its adapted LoRA decoder. We continue writing the remaining sessions with the retained source gate and evaluate the final memory using the target backbone. Previously processed sessions are not replayed, and the gate is not retrained. A reference run uses the target backbone's trained gate $G_B$ throughout the history.

\begin{table}[htbp]
\centering
\caption{Memory updates continued after backbone replacement on PERMA Noisy SD (accuracy, \%).}
\label{tab:additional-backbone-continuation}
\small
\begin{tabular}{lrr}
\hline
Backbone transition & Retained source gate $G_A$ & Target-trained gate $G_B$ \\
\hline
Qwen3-4B $\rightarrow$ Qwen3-8B & \textbf{85.37} & 91.37 \\
Qwen3-8B $\rightarrow$ Qwen3-4B & \textbf{76.02} & 86.00 \\
\hline
\end{tabular}
\end{table}

Retaining the source-trained gate produces 85.37\% accuracy after switching from 4B to 8B and 76.02\% in the reverse direction. Target-trained gate references achieve 91.37\% and 86.00\%, respectively. The source-gate runs demonstrate that accumulated memory and its existing consolidation policy can be preserved across backbone replacement, followed by additional session writes and memory-based prediction on the target backbone, with only the decoder requiring adaptation.
\par\endgroup

\FloatBarrier
\section{Additional Comparisons with Metis}
\label{app:metis-comparisons}

\subsection{Comparison on Qwen3.5-9B}
\label{app:same-backbone-comparison}

To control backbone family and scale in the comparison with Metis, we
evaluate No Context, Full Context, Metis-9B, and \methodname{} on Qwen3.5-9B.
All four methods share the \permabench{} task splits, visible-history boundary,
non-thinking mode, and A--H next-token-logit answer selection.
\methodname{} uses the adapted decoder described in
\appref{app:backbone-transfer} and fits its consolidation gate on the nine
training users of each fold.  Metis-9B uses its released memory-model weights
without additional \permabench{} training.

\begin{table}[htbp]
    \centering
    \caption{Same-backbone comparison on \permabench{} with Qwen3.5-9B
    (accuracy, \%).}
    \label{tab:same-backbone-comparison}
    \small
    \setlength{\tabcolsep}{4pt}
    \begin{tabular}{lrrrrrr}
        \toprule
        Method & Clean SD & Noisy SD & Clean MD & Noisy MD & Core Avg. & 7-Variant Avg. \\
        \midrule
        No Context & 29.20 & 29.20 & 35.63 & 35.63 & 32.41 & 31.98 \\
        Full Context & 69.27 & 72.00 & 85.71 & 85.68 & 78.17 & 74.25 \\
        Metis-9B & 80.87 & 80.47 & 80.17 & 79.29 & 80.20 & 80.04 \\
        \textbf{\methodname{} (ours)} & \textbf{96.50} & \textbf{96.50} &
        \textbf{92.52} & \textbf{92.68} & \textbf{94.55} & \textbf{94.66} \\
        \bottomrule
    \end{tabular}
\end{table}

Core Avg. averages the four displayed settings; 7-Variant Avg. additionally
includes Style SD, Style-Long SD, and Style MD.  Under this shared backbone
and evaluation protocol, \methodname{} exceeds Metis-9B by 14.35 pp on the
core average and 14.62 pp across all seven variants.

\subsection{Metis Scale Results}
\label{app:metis-scale}

We evaluate all three released Metis scales under the same seven-variant
\permabench{} protocol.
\autoref{tab:metis-scale} shows that Metis-9B obtains the highest overall
average and is therefore used in the main comparison; Metis-27B is strongest
only on Clean MD.

\begin{table}[htbp]
    \centering
    \caption{\permabench{} accuracy (\%) across released Metis model scales.}
    \label{tab:metis-scale}
    \small
    \resizebox{\linewidth}{!}{%
    \begin{tabular}{lrrrrrrrr}
        \toprule
        Model & Clean SD & Noisy SD & Style SD & Style-Long SD & Clean MD & Noisy MD & Style MD & Core Avg. \\
        \midrule
        Metis-4B & 73.77 & 74.67 & 73.27 & 73.13 & 70.81 & 70.71 & 68.34 & 72.49 \\
        \textbf{Metis-9B} & \textbf{80.87} & \textbf{80.47} & \textbf{79.67} &
        \textbf{79.97} & 80.17 & \textbf{79.29} & \textbf{79.87} & \textbf{80.20} \\
        Metis-27B & 78.67 & 74.87 & 77.03 & 77.03 & \textbf{81.55} & 78.45 & 78.65 & 78.39 \\
        \bottomrule
    \end{tabular}}
\end{table}

\FloatBarrier
\begingroup
\setlength{\intextsep}{16pt plus 2pt minus 2pt}
\setlength{\belowcaptionskip}{6pt}
\renewcommand{\arraystretch}{1.3}
\subsection{Target-Task Adaptation of the Metis Memory Writer}
\label{sec:additional-metis-writer}

We compare RPMem with a target-adapted Metis-9B Writer on PERMA Noisy SD. Both methods use Qwen3.5-9B-family backbones, ten leave-one-user-out splits, five training epochs, answer-label supervision, and the same 705 evaluation questions. For Metis we train its native memory Writer parameters across all eight memory blocks and rank-5 LoRA on writing-side key/value projections. The backbone and memory Reader are frozen; gradients from answer supervision propagate through the Reader to the Writer. This adaptation has 507,920 trainable parameters, compared with 524,800 parameters in RPMem's recurrent consolidation gate.

\begin{table}[htbp]
\centering
\caption{Target-task adaptation comparison on PERMA Noisy SD (accuracy, \%).}
\label{tab:additional-metis-writer}
\small
\begin{tabular}{lrrr}
\hline
Method & Adapted component & Trainable parameters & Accuracy \\
\hline
Metis-9B & None (released checkpoint) & 0 & 80.47 \\
Metis-9B & Memory Writer & 507920 & 87.00 \\
RPMem & Recurrent consolidation gate & 524800 & \textbf{96.50} \\
\hline
\end{tabular}
\end{table}

Target-task adaptation increases Metis accuracy from 80.47\% to 87.00\%, a gain of 6.53 percentage points. RPMem reaches 96.50\%, remaining 9.50 points higher under matched downstream supervision, evaluation, and training epochs and comparable trainable parameter budgets.

\subsection{Additional Comparison on LoCoMo (Gold)}
\label{sec:additional-locomo}

LoCoMo focuses on evidence-grounded factual question answering, including multi-hop and temporal questions. In contrast, the primary RPMem evaluation emphasizes preference formation and revision across sessions, changes in user information, and retention after intervening conversations. The Metis training corpus also includes synthetic examples seeded by entities extracted from LoCoMo, although its original question-answering behavior was not used. We include an additional evaluation using the LoCoMo (Gold) evidence-session setting reported by Metis.

\begin{table}[H]
\centering
\caption{LoCoMo (Gold) comparison (accuracy, \%). RPMem is evaluated in our additional experiment; the remaining values are reported by Metis (Table 6).}
\label{tab:additional-locomo}
\small
\begin{tabular}{llr}
\hline
Method & Memory access & Accuracy \\
\hline
RPMem (Qwen3.5-9B) & Parametric memory & 16.99 \\
Metis-9B & Parametric memory & 16.81 \\
RAG (top-5) & Retrieved text & 21.97 \\
Full Context & Gold-session context & \textbf{63.00} \\
\hline
\end{tabular}
\end{table}

RPMem reaches 16.99\%, compared with 16.81\% for Metis-9B, while RAG (21.97\%) and Full Context (63.00\%) perform better. This is consistent with the analysis in Metis~\citep{zhang2026metis}: direct access to original evidence benefits factual question answering (Section 6.2), whereas compression into a fixed-size latent memory state can lose information and confuse semantically similar facts (Section 8). These observations highlight the complementary value of explicit textual evidence for factual questions.
\par\endgroup

\FloatBarrier
\section{Evaluation Protocol Checks and Supplementary Diagnostics}
\label{app:memory-inversion}

We check prompt reproduction and answer scoring to contextualize the
\permabench{} comparisons.  Under the original prompt, Qwen3-32B reproduces
the official standalone Type-3 results within 0.7 pp in each setting;
\autoref{tab:protocol-reproduction} also reports the results under our prompt.
For Qwen3-8B in the No Context condition, API generation and local A--H logits
differ by 0.22 pp on SD and 1.90 pp on MD.

\begin{table}[H]
    \centering
    \setlength{\belowcaptionskip}{6pt}
    \renewcommand{\arraystretch}{1.3}
    \caption{Reproduction of \permabench{} standalone Type-3 accuracy (\%) with
    Qwen3-32B.}
    \label{tab:protocol-reproduction}
    \small
    \begin{tabular}{lccc}
        \toprule
        Setting & Reported & Official prompt & Our prompt \\
        \midrule
        Clean SD & 87.00 & 87.00 & 87.00 \\
        Noisy SD & 87.70 & 87.23 & 87.71 \\
        Clean MD & 93.60 & 93.63 & 87.90 \\
        Noisy MD & 93.00 & 92.36 & 89.17 \\
        Macro average & 90.33 & 90.05 & 87.94 \\
        \bottomrule
    \end{tabular}
\end{table}

Query-only evaluation provides a reference for performance available without
historical information.  Qwen3-32B reaches 75.58--75.89\% on Type~3 under this
condition, and Qwen3-8B reaches 52.70--66.95\% depending on domain.
\autoref{tab:gpt4o-memory-inversion} supplements these checks with our
GPT-4o-mini No Context reproduction and memory-system results reported by
\permabench~\citep{liu2026perma}; the table distinguishes their provenance.
Together, these results document protocol sensitivity and query-only
performance.  Attribution of the observed differences to information loss,
contextual interference, or answer-option cues requires additional controlled
comparisons.

\begin{table}[htbp]
    \centering
    \setlength{\belowcaptionskip}{6pt}
    \renewcommand{\arraystretch}{1.3}
    \caption{Supplementary GPT-4o-mini results on \permabench{} (accuracy, \%).
    No Context is our reproduction; all other rows are
    reported results from the benchmark paper.}
    \label{tab:gpt4o-memory-inversion}
    \small
    \begin{tabular}{lrrrrr}
        \toprule
        Method & Clean SD & Noisy SD & Clean MD & Noisy MD & Avg. \\
        \midrule
        \textbf{No Context} & 73.19 & 73.19 & \textbf{82.26} & \textbf{82.26} & 77.73 \\
        Full history & 78.00 & 76.60 & 70.70 & 72.00 & 74.33 \\
        RAG (BGE-M3) & 70.20 & 71.90 & 68.20 & 66.90 & 69.30 \\
        \textbf{MemOS} & \textbf{81.10} & \textbf{85.30} & 73.20 & 75.20 & \textbf{78.70} \\
        Mem0 & 68.60 & 66.00 & 65.00 & 66.20 & 66.45 \\
        LightMem & 65.70 & 67.10 & 60.50 & 63.10 & 64.10 \\
        Memobase & 73.30 & 68.30 & 69.40 & 64.30 & 68.83 \\
        EverMemOS & 72.80 & 69.50 & 71.30 & 73.20 & 71.70 \\
        Supermemory & 65.50 & 67.40 & 65.60 & 63.70 & 65.55 \\
        \bottomrule
    \end{tabular}
\end{table}

\FloatBarrier
\begingroup
\setlength{\belowcaptionskip}{6pt}
\renewcommand{\arraystretch}{1.3}
\subsection{Relevant-Context Analysis for Qwen3.5-35B-A3B}
\label{sec:additional-relevant-context}

We further examine the degradation of Qwen3.5-35B-A3B under Full Context on PERMA. The Relevant Context condition supplies only task-relevant historical information. The other three results are from the original cross-backbone evaluation.

\begin{table}[H]
\centering
\caption{Qwen3.5-35B-A3B performance on PERMA (accuracy, \%).}
\label{tab:additional-relevant-context}
\small
\begin{tabular}{lr}
\hline
Context condition & Accuracy \\
\hline
No Context & 79.48 \\
Full Context & 64.86 \\
Relevant Context & 83.78 \\
RPMem & \textbf{94.21} \\
\hline
\end{tabular}
\end{table}

Full Context trails No Context by 14.62 percentage points. Relevant Context reaches 83.78\%, 4.30 points above No Context. This pattern is consistent with interference from irrelevant historical information when the model must identify and use task-relevant evidence from a long input context.
\par\endgroup

\FloatBarrier
\section{Additional Gate Dynamics Visualization and Case Studies}
\label{app:gate-cases}

\begin{figure}[htbp]
    \centering
    \includegraphics[width=0.95\linewidth]{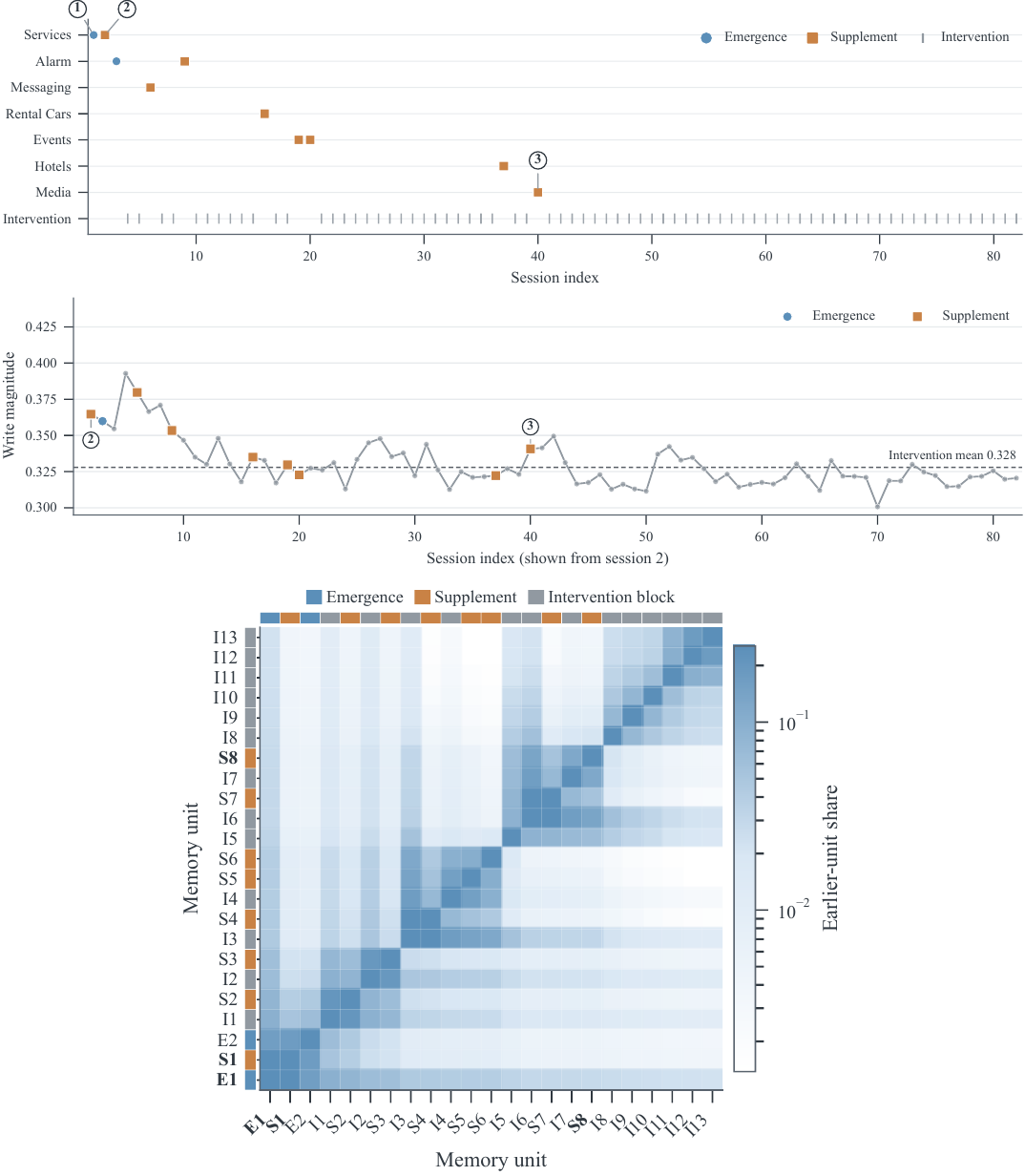}
    \caption{Representative held-out trajectory: event timeline (top),
    session-wise write magnitude from session 2 onward (middle), and
    memory-unit source shares (bottom). The timeline and heatmap cover the
    complete history. E and S denote emergence and supplement events on the task's preference
    timeline; I denotes a block of consecutive intervention sessions.
    An intervention may replay a preference event from another timeline,
    including a domain's emergence. For each pair of
    units, the heatmap gives the earlier unit's summed source share at the
    end of the later unit, mirrored across the diagonal. Color intensity uses
    a logarithmic scale. Circled numbers identify three selected events.}
    \label{fig:gate-case}
\end{figure}

\autoref{fig:gate-case} presents a deterministically selected representative
trajectory: among 423 candidates with at least 40 sessions, three domains, and
both emergence and supplement events, it is nearest to the multivariate median
of four gate statistics.
The trajectory contains 82 sessions for held-out user 507, including ten
preference events from seven domains and 72 distractor sessions.
The write curve displays sessions 2--82, with the first session omitted.
This instance illustrates the aggregate pattern in
\autoref{sec:source-survival}: early domain-establishing events can form a
long-lived backbone even when later same-domain updates temporarily dominate.

\FloatBarrier
\begingroup
\subsection{Controlled Preference Intervention in PrefEval}
\label{sec:additional-functional-case}

We study whether a controlled change to a specific preference affects subsequent model predictions while an independent preference remains accessible. We use Qwen3-8B with the trained PrefEval memory compiler (fixed-reference forward-KL checkpoint at 51,885 updates) and the five-epoch recurrent consolidation gate. All model parameters remain frozen. Two histories differ only in one target preference concerning Python learning materials. Both histories contain the same independent car-rental preference (the user drives only manual-transmission cars), followed by the same ten intervening user--assistant dialogue turns (20 messages). The two histories are independently compiled and consolidated from the same initial memory state.

The intervened preference in History A states: ``I avoid educational content that is not accessible to the visually impaired. When I choose learning materials, I need to be able to use them without depending on clear sight. A resource that relies on small printed text, visual demonstrations or inspecting a screen would create an accessibility barrier for me. Please prioritize ways of learning that remain usable without those visual demands. Accessibility is essential to my participation, not an optional extra.''

History B instead states: ``I strongly dislike video and audio-based educational resources. I prefer written content and textbooks because I can read at my own pace, revisit a passage and organize my notes around the text. Listening to a recording or following a video does not suit the way I study. When I choose learning material, I want a text-based resource with written explanations I can work through, rather than an audio or video lesson.''

After ten intervening turns, we ask two separate multiple-choice questions. The target question is ``What are the best resources to learn Python programming?'' with candidate responses consisting of video tutorials, interactive visual coding challenges, audiobooks for visually impaired learners, and printed Python books. The independent control question requests a reliable rental car for an upcoming trip to Miami, with the manual-transmission Volkswagen Golf TSI among the options. During evaluation, the reader receives only the question, the four choices, and an answer-format instruction. Historical information is accessed through the consolidated memory decoded into LoRA parameters, without adding the past conversation as text.

\begin{table}[htbp]
\centering
\caption{Predictions after ten identical intervening dialogue turns in paired PrefEval histories. The native and reversed conditions use different answer-option orders.}
\label{tab:additional-functional-case}
\small
\begin{tabular}{lcc}
\hline
History and option order & Python learning (target) & Car rental (control) \\
\hline
A, native & Audiobooks (C) & Manual Golf (B) \\
B, native & Printed books (D) & Manual Golf (B) \\
A, reversed & Audiobooks (B) & Manual Golf (C) \\
B, reversed & Printed books (A) & Manual Golf (C) \\
\hline
\end{tabular}
\end{table}

Replacing the learning preference changes the target answer from audiobooks to printed books under both answer-option orders, while the independent car-rental answer remains unchanged. The case illustrates that specific information carried through the recurrent memory affects later predictions after additional session updates, complementing the coefficient-based memory-dynamics analysis.
\par\endgroup

\end{document}